\documentclass[journal,compsoc,cspaper,final]{IEEEtran}
\usepackage{ragged2e}
\usepackage[export]{adjustbox}
\usepackage{tikz}
\usetikzlibrary{spy}
\usepackage{comment}
\usepackage{amsmath,amssymb} % define this before the line numbering.
\usepackage{color}
\usepackage{makecell}
\usepackage{multirow}
\usepackage[pagebackref,breaklinks,colorlinks]{hyperref}
\usepackage[keeplastbox]{flushend}
\usepackage{forloop}
\usepackage[capitalize]{cleveref}
\crefname{section}{Sec.}{Secs.}
\Crefname{section}{Section}{Sections}
\Crefname{table}{Table}{Tables}
\crefname{table}{Tab.}{Tabs.}

\usepackage{subcaption}
\usepackage{animate}
\usepackage{xspace}
\usepackage{pifont}
\usetikzlibrary{fit}

\makeatletter
\DeclareRobustCommand\onedot{\futurelet\@let@token\@onedot}
\def\@onedot{\ifx\@let@token.\else.\null\fi\xspace}

\def\eg{\emph{e.g}\onedot} 
\def\ie{\emph{i.e}\onedot} 
 
\def\etc{\emph{etc}\onedot} 
 
\def\etal{\emph{et al}\onedot}
\makeatother

\ifCLASSOPTIONcompsoc
  \usepackage[nocompress]{cite}
\else
  \usepackage{cite}
\fi
\ifCLASSINFOpdf
\else
\fi
\begin{document}
% \title{Demystifying Single Blurry Image with Events}
% \title{Motion Deblurring in Single Images with Events}
% \title{Demystifying the Motion Blurs \\ in Single Blurry Images with Events}
% \title{Demystifying the Motion Blurs }
% \title{Continuous-time Video Extraction from a Single Blurry Image with Events}
\title{Recovering Continuous Scene Dynamics from \\A Single Blurry Image with Events}

\author{Zhangyi Cheng, Xiang Zhang,
        Lei Yu, Jianzhuang Liu, Wen Yang,
        and~Gui-Song Xia% <-this % stops a space
        \IEEEcompsocitemizethanks{
\IEEEcompsocthanksitem Z. Cheng and G.-S. Xia are with the School of Computer Science, Wuhan University, Wuhan 430072, China. \\E-mail: \{zyc,guisong.xia\}@whu.edu.cn.
\IEEEcompsocthanksitem X. Zhang, L. Yu, and W. Yang are with the School of Electronic Information, Wuhan University, Wuhan 430072, China. \\ E-mail: \{xiangz, ly.wd, yangwen\}@whu.edu.cn.
\IEEEcompsocthanksitem J. Liu is with the Huawei Noah’s Ark Lab, Shenzhen 518000, China. \\ E-mail: liu.jianzhuang@huawei.com.
\IEEEcompsocthanksitem The research was partially supported by the National Natural Science Foundation of China under Grants 62271354, 61871297, 61922065, 41820104006, 61871299, and the Natural Science Foundation of Hubei Province, China under Grant 2021CFB467.
\IEEEcompsocthanksitem Z. Cheng and X. Zhang contributed equally to this work.
\IEEEcompsocthanksitem Corresponding authors: L. Yu and G.-S. Xia.
}
}


\markboth{TPAMI Submission}
% \markboth{Journal of \LaTeX\ Class Files,~Vol.~14, No.~8, August~2015}%
{Shell \MakeLowercase{\textit{et al.}}: Bare Demo of IEEEtran.cls for Computer Society Journals}



\IEEEtitleabstractindextext{%
\begin{abstract}
\justifying
This paper aims at demystifying a single motion-blurred image with events and revealing temporally continuous scene dynamics encrypted behind motion blurs. To achieve this end, an Implicit Video Function (IVF) is learned to represent a single motion-blurred image with concurrent events, enabling the latent sharp image restoration of arbitrary timestamps in the range of imaging exposures. Specifically, a dual attention transformer is proposed to efficiently leverage merits from both modalities, \ie, the high temporal resolution of event features and the smoothness of image features, alleviating temporal ambiguities while suppressing the event noise. The proposed network is trained only with the supervision of ground-truth images of limited referenced timestamps. Motion- and texture-guided supervisions are employed simultaneously to enhance restorations of the non-referenced timestamps and improve the overall sharpness. Experiments on synthetic, semi-synthetic, and real-world datasets demonstrate that our proposed method outperforms state-of-the-art methods by a large margin in terms of both objective PSNR and SSIM measurements and subjective evaluations. 
%vents. Codes and datasets are available at \url{http://dvs-whu.cn/}.
\end{abstract}

}

% \usepackage{widetext}

% \twocolumn[{
% \renewcommand\twocolumn[1][]{#1}
\maketitle
\IEEEdisplaynontitleabstractindextext

% \fi
\IEEEpeerreviewmaketitle

% \IEEEraisesectionheading{\section{Introduction}\label{sec:introduction}}

% \input{sec/1_introduction.tex}
% \input{sec/2_related.tex}
% \input{sec/3_method.tex}
% \input{sec/4_results}
% \input{sec/5_conclusions}
%%%%%%%%%%%%%%%%
% Introduction %
%%%%%%%%%%%%%%%%
\IEEEraisesectionheading{\section{Introduction}\label{sec:introduction}}
\label{sec:intro}
% 

% 1. motion blurred图像中包含texture和motion信息，recover single image，丢失了motion信息，从而导致sequence的temporal ambiguity。xxx 和 xxx 利用sophestic的方法来消除temporal的ambiguity，
% 2. event数据提供了高帧率的intra-frame motion information，因此利用event可以帮助我们去进一步explore motion blurred image中的信息。EDI、LEDVDI、RED-Net、eSLNet等研究了如何利用event从motion blurred的图像中重建图像的过程。

% \IEEEPARstart{P}{erction} with the emergence of motion blur is a nuisance 
% deserves to be demystified to perceive the scene dynamics behind the blurred photographs~\cite{hqdeblurring_siggraph2008}. 
% Motion deblurring targets at , especially when encountering complex motions or severely blurred frames~\cite{zhang2021exposure}
% it is extremely challenging to achieve CTVB due to motion ambiguities and texture erasures caused by motion blurs~\cite{gupta2010single}. Furthermore, 

\IEEEPARstart{M}{otion} blur is a nuisance that commonly exists in photographs when perceiving scenes with relative motions from the camera to the targets~\cite{hqdeblurring_siggraph2008}. Most existing motion deblurring approaches commonly focus on restoring a single image~\cite{krishnan2011blind,liu2020self,tai2011,chen2019d,xu2013unnatural,fergus2006removing,pan2016blind,gong2017motion,nimisha2017blur} or a discrete-time video sequence~\cite{zhang2020g,jin2018learning,jin2019learning,purohit2019bringing,rengarajan2020photosequencing} from the blurry input, unable to reveal continuous scene dynamics and provide every subtle moment behind the blurriness~\cite{zhang2021exposure}. 
Continuous-time video extraction from a single blurry image reveals the temporal continuous scene dynamics by restoring the sharp latent images of arbitrary timestamps in the range of the exposure time interval, benefiting many real-world applications, \eg, sports photography, industrial monitoring, image segmentation, target tracking, and object recognition. %Nevertheless, CTVB is still 

% we have to explore motions or textures in  in restoring sharp latent images of arbitrary timestamps in the range of exposure time intervals requires either continuous. However, it is extremely challenging to achieve CTVB since 

% Compared with the conventional motion deblurring tasks, CTVB is more challenging. 
% Conventional frame-based motion deblurring approaches mostly focus on restoring a single image or a discrete-time video sequence. 
The inversion of the blurry process is commonly ill-posed since real-world blurry images are temporal integrations of the continuous scene dynamics with missing information on intra-frame motions and textures~\cite{gupta2010single}. %Revealing temporal continuous scene dynamics from a single blurry image is challenging
Such ill-posedness can be relieved upon proper pre-defined priors/assumptions either on motions~\cite{krishnan2011blind,liu2020self,tai2011} or intensity textures~\cite{chen2019d,xu2013unnatural,fergus2006removing,pan2016blind}. Nevertheless, the performance of conventional approaches is confined 
to the above fragile assumptions. Even though the end-to-end learning networks supervised by paired datasets can significantly improve the deblurring performance~\cite{gong2017motion,nimisha2017blur} and even achieve sequence restoration by decoupling the temporal motion ambiguity~\cite{purohit2019bringing,jin2018learning,rengarajan2020photosequencing}, the lack of continuous-time representations in terms of the intra-frame motions or textures hinders most existing frame-based approaches being applied for continuous-time video extraction from a single blurry image. 

Learning a continuous-time representation of the intra-frame motions and textures from a single blurry image is generally difficult. Motion kernels~\cite{chen2019d,xu2013unnatural}, optical flows~\cite{chen2018reblur2deblur,liu2020self}, and exposure trajectories \cite{zhang2021exposure} are typical tools for continuous-time motion representations, which however often suffer from ill-posed nature of blur estimation. While the continuous-time texture representations are indirectly considered as the inter-frame consistencies between two consecutive blurry frames~\cite{jin2019learning}, it is still struggling to provide accurate estimations in terms of the missing motions and the erased textures during the whole exposure period, especially when encountering complex motions or severely blurred frames, and thus far from achieving continuous-time video restorations.

% To achieve CTVB, 

%  continuous time motion representation ~ Motion-ETR （Motion估计误差）
%  continuous time texture representation ~ E-CIR (polynomial引入一些误差，噪声会影响polynomial fitting，Motion缺失）、event-reshuffle（正反阈值会导致reveral的model与前向不同）
%  Motion + texture

% the frame-based methods have confined performance for general real-world scenarios, especially when encountering complex motions or severely blurred frames. 

% Nevertheless, the resulting algorithms largely rely on the domain-specific would suffer from performance degradation due to the violation of 
% existing frame-based methods a single \cite{pan2016blind,sun2013edge} or a sequence \cite{jin2018learning,purohit2019bringing,rengarajan2020photosequencing}

% to achieve the end 
% is an ill-posed problem due to the motion ambiguities and the texture erasures~\cite{gupta2010single}. Existing motion-deblurring approaches leverage artificial or data-driven priors to achieve the end but only a single \cite{pan2016blind,sun2013edge} or a sequence \cite{jin2018learning,purohit2019bringing,rengarajan2020photosequencing} of sharp latent images aligning to discrete timestamps can be restored, resulting in missing of crucial moments behind motion blurs, as shown in Fig.~\ref{fig:ivf_1}.

% Regarding the motions

% Regarding the textures

% Event-based approaches blabla ... 

\begin{figure*}[t!]
    \centering
    \includegraphics[width=\linewidth]{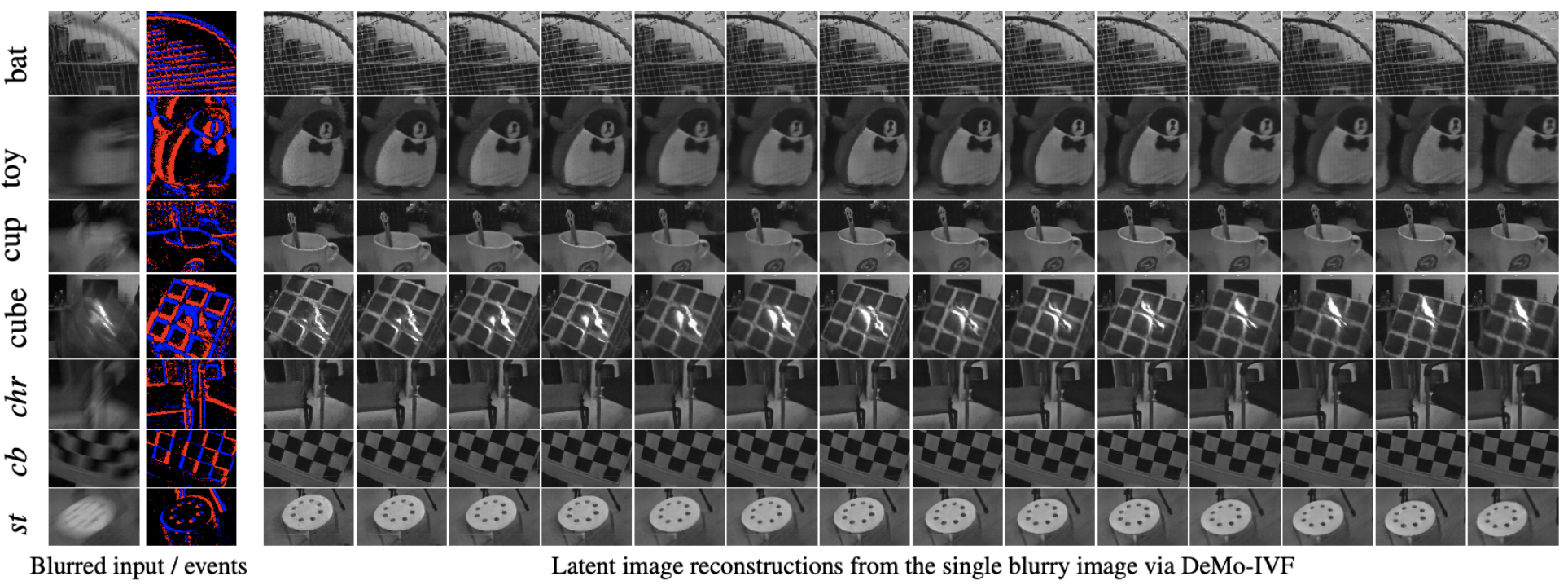}
 \caption{Qualitative results of our proposed DeMo-IVF on a real-world dataset. An Implicit Video Function (IVF) is learned to represent a single blurry image with concurrent events. Using IVF, we query $109$ latent images from a single blurry image of different scenes, where {\it chr, cb}, and {\it st} respectively denote {\it chair, chessboard}, and {\it stool}, and we select the first $14$ frames for visualization.}
	\label{fig:ivf_1}
 \end{figure*}

In this paper, we propose to introduce the event camera to alleviate the burdens of continuous-time video extraction which can restore latent sharp images of arbitrary timestamps as shown in Fig.~\ref{fig:ivf_1}. Different from the conventional frame-based cameras, event cameras perceive the scene dynamics by encoding brightness changes with extremely low latency (in the order of $\mu$s) and asynchronously emitting binary events with extremely high temporal resolution \cite{lichtsteiner128Times1282008,gallego2020event}. Thus events can provide intra-frame clues about motions \cite{zhu2018ev} and intensity textures \cite{rebecq2019c}, which bridges the gap between the blurry observations and the sharp latent images \cite{pan2019bringing}, unlocking the potential to reveal continuous-time scene dynamics behind blurry images. Many event-based motion deblurring approaches have recently been proposed and achieved prominent deblurring
performance even for complex motions or severely blurred frames~\cite{pan2020high,jiang2020learning,xu2021motion,wang2020event,zhang2022unifying,song2022cir,lin2020learning,shang2021bringing}. However, most existing works are dedicated to restoring a single image or a discrete-time video sequence instead of continuous-time video restorations.

% Events convey intra-frame information on both motions and textures in a nearly continuous form, 
Similar to frame-based approaches, the performance of restoring coninuous-time videos with events largely relies on the accuracy of continuous-time representations for the intra-frame motions and textures. 
% Recently, we have witnessed significant progress on event-based motion deblurring~\cite{pan2020high,jiang2020learning,xu2021motion,wang2020event,zhang2022unifying,song2022cir,lin2020learning,shang2021bringing}, but most existing approaches~\cite{jiang2020learning,xu2021motion,lin2020learning,shang2021bringing} are dedicated to restoring a single image or a discrete-time video sequence instead of CTVB. 
A sophisticated event reshuffle process~\cite{pan2020high,wang2020event,zhang2022unifying} is proposed as a time-dependent event representation to provide continuous-time texture compensations and finally achieve arbitrary-timestamp restorations. However, since the event thresholds are different between positive and negative polarities, the reshuffle process inevitably alternates event polarities and thus brings modeling errors~\cite{gallego2020event}. Meanwhile, parametric polynomials have been employed to approximate the per-pixel continuous-time intensity functions by fitting the temporal derivatives with events~\cite{song2022cir}, but the massive amount of event noise in spatial and temporal domains~\cite{inivaiton2020} would inevitably mislead the intensity polynomials, especially in static regions only with event noise. On the other hand, since instance optical flows would be predicted from events~\cite{zhu2018ev,wan2022learning}, we can fulfill the continuous-time video restoration by warping latent sharp restored images to any specific timestamps or turning to event-based video interpolation approaches~\cite{tulyakov2021time,tulyakov2022time,he2022timereplayer}. Nevertheless, the prediction error on optical flows and latent restorations can further be propagated to the final results. %\textcolor{red}{(ZX: how to validate the EV-Flow in our method does not suffer from the error propagation?)}

Inspired by the implicit neural representation~\cite{mildenhall2020c}, we first bridge the gap between discrete-time and continuous-time video representations by learning an Implicit Video Function (IVF) from blurry images and events. It is more challenging than existing works using sharp and clear video clips \cite{Bemana2020xfields,benbarka2021,chen2021b,sitzmann2020,tancik2020}.
% This paper bridges the gap between discrete-time and continuous-time video representations for motion deblurring to restore the latent images at arbitrary time instances as shown in Fig.~\ref{fig:ivf_1}. Inspired by the implicit neural representation \cite{mildenhall2020c}, we propose to learn an Implicit Video Function (IVF) from blurry images and the concurrent events, which is more challenging than existing works (using sharp and clear video clips) \cite{Bemana2020xfields,benbarka2021,chen2021b,sitzmann2020,tancik2020}. 
To ease the burden, the overall architecture of the proposed IVF is divided into the temporally constant components, \ie, the blurry image, and the temporally alternating component, \ie, a continuous-time function closely related to events.
% (as a compensation). Then a continuous-time function to represent the alternating component which is closely related to events. 
For event noise, we mutually compensate events and frames to enhance the performance of IVF since the blurry image is less noisy than events while events are immune to motion blurs. 
% us to learn the IVF by mutual compensations. 
We employ the conditioned Multi-Layer Perceptrons (MLP) with the Fourier position encoding scheme to learn IVF and a Dual Feature Embedding Network (DFEN) to take the merits from both events and the blurry input in the feature domain. 
To fully utilize the property of extremely high temporal resolution of events, the initial restorations from IVF are further refined with time-dependent subtle event segments through an Event-based Edge Refinement (EER) module.
The overall network is only supervised by ground-truth images
of limited referenced timestamps. We further introduce motion- and texture-guided supervisions to enhance restorations of the non-referenced timestamps and improve the overall sharpness. 

The contributions of this paper are three-fold:
% \vspace{-1em}
\begin{itemize}
\item We propose to learn an Implicit Video Function from blurry images and the concurrent events, which fully {\it De}mystifies {\it Mo}tion blurred images (DeMo-IVF) and produces temporally continuous sharp sequences. To the best of our knowledge, this is the first work about learning an implicit video function from a single blurry frame and concurrent events. 
% Finally, we propose to fully {\it De}mystify {\it Mo}tion blurred images (DeMo) based on the learned IVF, \ie, DeMo.
% \vspace{-1em}
\item We propose a dual feature embedding network, \ie, DFEN, to simultaneously consider the event noise suppression and the blurry feature enhancement by multi-stage transformers. Meanwhile, an Event-based Edge Refinement (EER) module is also presented to enhance the overall texture restoration performance.

% we propose a two-stage training strategy to supervise the proposed network where motion- and texture-guided supervisions are presented to enhance the overall texture restoration performance.
% \vspace{-1em}
\item We propose to train the overall network with motion- and texture-guided supervisions only based on ground-truth images of limited referenced timestamps. We evaluate our proposed DeMo-IVF on synthetic, semi-synthetic, and real-world datasets, showing that DeMo-IVF outperforms state-of-the-art methods in both restoration quality and temporal resolution. 
% Ideally, the proposed DeMo can restore $\infty$ frames from a single blurry image, thus fully demystifying motion blurs.
\end{itemize}

% In a nutshell, contributions of this paper are three-fold:
% % \vspace{-1em}
% \begin{itemize}
% % \setlength\itemsep{-.3em}
% \item We propose to learn an Implicit Video Function (IVF) from blurry images and concurrent events. To the best of our knowledge, this is the first work about learning an IVF from blurry and noisy inputs. Finally, we propose to fully {\it De}mystify {\it Mo}tion blurred images (DeMo) based on the learned IVF, \ie, DeMo.
% % \vspace{-1em}
% \item We propose a Dual Feature Embedding Network (DFEN) to simultaneously consider the event noise suppression and the blurry feature enhancement by multi-stage transformers.
% % \vspace{-1em}
% \item We evaluate the DeMo on both synthetic and real-world datasets, showing that DeMo outperforms state-of-the-art methods in restoration quality and temporal resolution. Ideally, the proposed DeMo can restore $\infty$ frames from a single blurry image, thus fully demystifying motion blurs.
% \end{itemize}

%%%%%%%%%%%%%%%%
% Related Work %
%%%%%%%%%%%%%%%%
\section{Related Work}
\label{sec:related}

% \AT{introduce subtopics, mention surveys here}
% \lorem{1}

 %< bloody latex and its heuristics for figure placement

% \paragraph{Motion Deblurring}
% Motion blur commonly happens with moving objects during the exposure period, leading to visual degeneration and inherently encodes motions and textures in captured blurry images \cite{jin2018learning}. 
\noindent \textbf{Frame-based Motion Deblurring.}
The task of motion deblurring aims to restore sharp clear latent images and reveal the hidden information behind motion blurs which, however, is generally ill-posed. We can roughly categorize existing approaches into {\it single image} and {\it video sequence} according to the number of restored image frames.
For the single image restoration,  early attempts of optimization-based methods generally require properly pre-defined priors/assumptions either on motions or on intensity textures; \eg, linearity \cite{liu2020self} and projective motion path \cite{tai2011} are often assumed for motions, and priors like gradient prior \cite{chen2019d}, sparsity \cite{xu2013unnatural,hu2010}, Gaussian scale mixture \cite{fergus2006removing} and dark channel \cite{pan2016blind} are exploited for intensity textures. 
However, identifying a suitably informative and general prior is difficult and crucial for the deblurring performance, and improper priors may lead to artifacts and degraded results \cite{zhou2019spatio}.
% However, identifying a suitably informative and general prior for motion blurs is struggling, while improper prior may even generate artifacts and degrade deblurring performance \cite{zhou2019spatio}.
To overcome the limitations of manual priors, learning-based methods leverage the merits of convolutional neural networks (CNNs) to predict latent sharp images in an end-to-end manner, achieving prominent performance \cite{gong2017motion,nimisha2017blur}. However, single-image restoration reveals only the static intensity textures but loses the entire motion information behind the blurry image.  
% scene dynamics, which is an important message behind motion blurs.

% The {\bf $\mathbf{1}$-to-many} motion deblurring methods restore a video sequence and reveal the scene dynamics, 
Existing video sequence restoration methods reveal the scene dynamics by extracting a predefined fixed number of latent images from one blurry frame, where the temporal ambiguity is one of the critical challenges \cite{zhang2020g,jin2018learning,jin2019learning,purohit2019bringing,rengarajan2020photosequencing}. Constraints of temporal ordering \cite{jin2018learning,purohit2019bringing} and motion consistency \cite{zhang2020g} have been investigated to decouple such temporal ambiguity. On the other hand, the inter-frame consistency has also been exploited to learn time ordering from two consecutive blurry images \cite{jin2019learning} or clear images (with short exposure time but noisy) \cite{rengarajan2020photosequencing}. 
% However, to the best of our knowledge, existing approaches only focus on the task of motion deblurring with a fixed number of reconstructions, which cannot be directly employed to restore the latent sharp images of arbitrary timestamps, resulting in the missing of crucial moments \cite{rozumnyia}. (\textcolor{red}{contradict with sec 5.2})
Despite these efforts, the image-only approaches still struggle in sequence restoration, especially when the motion blur is large. 
Without additional auxiliary signal input, arbitrary frame reconstruction is even more unattainable.

\noindent \textbf{Event-based Motion Deblurring.}
Benefiting from the extremely high temporal resolution, events can provide the missing intra-frame information about motions and intensity textures \cite{lichtsteiner128Times1282008}. Hence, the gap between the blurry observations and the latent sharp images can be potentially bridged \cite{pan2019bringing}. According to the physical model of event cameras \cite{gallego2020event}, explicit relations are built between events and images. Many continuous-time event-based motion deblurring algorithms are proposed such as the complementary filter \cite{scheerlinck2018continuous}, the event-based double integral (EDI) \cite{pan2019bringing}, the asynchronous spatial convolution \cite{Scheerlinck19ral}, and the asynchronous Kalman filter \cite{Wang21iccv}.
% As pioneer efforts, the complementary filter \cite{scheerlinck2018continuous}, the event-based double integral (EDI) \cite{pan2019bringing}, the asynchronous spatial convolution \cite{Scheerlinck19ral} and the asynchronous Kalman filter \cite{Wang21iccv} have been proposed for motion deblurring and high frame rate video reconstruction. According to the physical model of event cameras \cite{gallego2020event}, explicit relations are built between events and images resulting in continuous-time event-based motion deblurring algorithms. 
However, the ideal model of event generation is often disturbed in real-world scenarios due to the huge amount of noise caused by the imperfection of physical circuits \cite{inivaiton2020}, leading to performance degradation in practice \cite{scheerlinck2018continuous}.

% Compared to these model-based approaches, learning-based methods effectively suppress noise by fitting the distribution of noisy events \cite{wang2020event,wang2020joint,lin2020learning,xu2021motion}. eSL-Net is proposed to deal with event noise by leveraging the sparsity behind latent images \cite{wang2020event}, but it may result in over-smoothness due to the inconsistency between the synthetic events (for training) and the real-world events (for inference). LEDVDI tackles such inconsistency by training directly on real-world events \cite{lin2020learning}, and RED-Net further improves the deblurring performance by training the network with both synthetic and real-world events in a semi-supervised manner \cite{xu2021motion}. However, existing learning-based approaches with events only focus on restoring the latent images at pre-defined timestamps and a re-training phase or a cascaded interpolation network is required to enable reconstruction at arbitrary time instances.
Compared to these model-based approaches, learning-based approaches effectively suppress noise by fitting the distribution of noisy events \cite{wang2020event,wang2020joint,lin2020learning,xu2021motion}. 
Jiang \etal \cite{jiang2020learning} reinterpret a sequential deblurring process by a convolutional recurrent neural network.
Lin \etal \cite{lin2020learning} use events to estimate the residuals of deblurring and interpolation for sharp frame restoration, and they propose to use a dynamic filtering layer to handle spatially varying triggering thresholds for events.  
Xu \etal \cite{xu2021motion} exploit photometric consistency and blurry consistency to train the network with both synthetic and real-world data in a semi-supervised manner to bridge the synthesis-to-reality gap.
% between simulated and real-world motion blurs. 
% D2Net
Shang \etal \cite{shang2021bringing} assume that sharp frames usually appear nearby blurry frames and propose a framework for tackling video deblurring with non-consecutive blurry frames.
% EVDI
However, the aforementioned event-based deblurring neural networks only focus on restoring the latent images at pre-defined timestamps and a re-training phase or a cascaded interpolation algorithm is required to enable reconstruction at arbitrary timestamps. To achieve the continuous-time video restoration, the event re-shuffle process is employed in eSL-Net~\cite{wang2020event} and EVDI~\cite{zhang2022unifying}, implemented by splitting events and then reversing their temporal orders and polarities. But the event re-shuffle process would introduce modeling error induced by reversing polarities since positive and negative events are commonly triggered with different contrast thresholds~\cite{brandli2014b}. Meanwhile, E-CIR~\cite{song2022cir} queries latent images of arbitrary timestamps by fitting the per-pixel parametric polynomials with events but often suffers from noise artifacts.

\vspace{1em}

Therefore, learning an effective continuous-time video representation is important to recover continuous scene dynamics. To achieve this end, we borrow the idea from the Implicit Neural Representation (INR) \cite{mildenhall2020c}, which approximates the continuous functions that map the domain of the input signal (coordinates, time, voxel, \etc) to a representation of color, amplitude, or density at an arbitrary input location~\cite{benbarka2021,chen2021b,sitzmann2020,tancik2020}. The idea of learning the INR with multi-layer perceptrons has been widely applied in various fields such as 3D rendering \cite{mildenhall2020c}, video generation \cite{niemeyer2019}, and image representation \cite{chen2021b}. Considering the motion deblurring task with continuous time instances, we cast it as learning an implicit video function (IVF) from a blurry image. Unlike existing approaches \cite{niemeyer2019,mildenhall2020c,pumarola2020,tretschk2021} that aim to learn an IVF from multi-view sharp and clear images, our task directly accepts motion-blurred input which is more challenging.

\begin{figure*}[!th]
\centering
\includegraphics[width=1\textwidth]{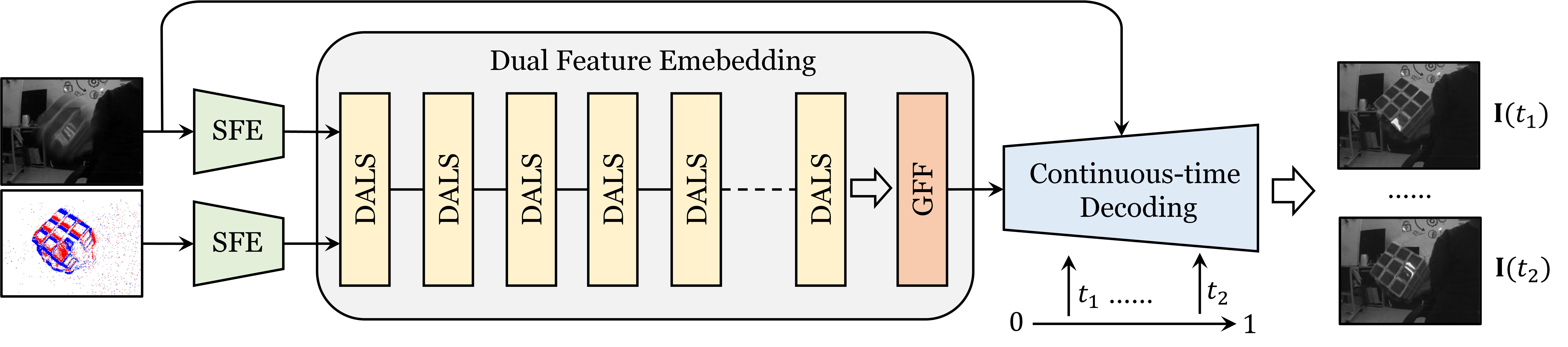}
\caption{Architecture of the proposed IVF, which is composed of Dual Feature Embedding Network (DFEN) and Continuous-time Decoding MLP. A single blurry image and the concurrent events are separately processed in DFEN by multi-layer transformers with the dual attention mechanism for event noise suppression and blurry feature enhancement. Then the latent sharp images at arbitrary timestamps $t$ are restored by the Continuous-time Decoding MLP. }
%\textcolor{red}{(ZX: do we need to depict EER in this figure?)}}
\label{fig:overview_dals}
\end{figure*}
% \section{Problem Statement}
\section{Method}

\subsection{Problem Formulation}
Physically, a motion-blurred image $\bf B ({\bf x})$ can be expressed as the average of latent images over the exposure period $\mathcal{T}$,
\begin{equation}\label{eq:blur}
{\bf B} ({\bf x}) = \frac{1}{|\mathcal{T}|}\int_{t\in \mathcal{T}} {\bf I}({\bf x},t) dt ,
\end{equation}
where the latent scene dynamics ${\bf I}({\bf x},t)$ is a continuous function mapping positions $\bf x$ and timestamps $t$ to pixel values\footnote{We drop $\bf x$ for simplification in the following.}. 
Given a finite index set $\mathcal{N} \subset \mathbb{Z}$, restoring the $n$-th latent sharp image ${\bf I}_n, n\in \mathcal{N},$ of the continuous scene dynamics ${\bf I}(t)$ from the blurry image $\bf B$ is an ill-posed problem \cite{jin2018learning,purohit2019bringing,rengarajan2020photosequencing}.
Many algorithms have been proposed but only for the restoration of discrete version of ${\bf I}(t)$, \ie, ${\bf I}_n$ corresponding to the latent clear image at time $t_n\in \mathcal{T}$, through event-based motion deblurring networks (denoted as EMD-Net)\cite{lin2020learning,jiang2020learning,wang2020event,xu2021motion}.
\begin{equation}\label{eq:deblurNet}
    \{{\bf I}_n\} = \mbox{EMD-Net}({\bf B}, \mathcal{E}_{\mathcal{T}}),
\end{equation} 
where $\mathcal{E}_{\mathcal{T}} \triangleq \{({\bf x}_i,p_i,t_i)\}_{t_i\in \mathcal{T}}$ is the set of events triggered in $\mathcal{T}$ with $t_i$ and ${\bf x}_i$ respectively denoting the timestamp and the pixel location of the $i$-th event, and $p_i\in \{+1,-1\}$ denoting the polarity. $\text{EMD-Net}(\cdot)$ is a sequence reconstruction operator, and once it has been trained, only the latent clear images of fixed timestamps $t_n$ can be predicted. Thus, an interpolation algorithm is often required to restore the latent images off the fixed timestamps \cite{lin2020learning}, which increases the complexity and even results in sub-optimal solutions due to the propagation of deblurring errors.

In this paper, we target fully {\it De}mystifying {\it Mo}tion-blurred images (DeMo) with events by directly recovering the temporally continuous scene dynamics ${\bf I}(t)$ behind motion blurs. 
Different from Eq. \eqref{eq:deblurNet}, the task of DeMo is to restore sharp latent images of any timestamps $t$ during the exposure time interval $\mathcal{T}$. It can be realized by learning an implicit neural representation, \ie,
%\cite{niemeyer2019,mildenhall2020c,pumarola2020,tretschk2021}
\begin{equation}\label{eq:demo}
    {\bf I}(t) = \mbox{DeMo}(t; {\bf B}, \mathcal{E}_{\mathcal{T}}), \forall t\in \mathcal{T},
\end{equation}
which is an implicit video function of time $t$ conditioned on a single blurry image $\bf B$ and the corresponding event stream $\mathcal{E}_{\mathcal{T}}$. Different from achieving the DeMo task in two stages, \ie, deblurring and temporal upsampling \cite{jin2018learning,purohit2019bringing}, we aim at learning the scene dynamics as an Implicit Video Function (IVF) from a blurry image and the concurrent events. The resulting IVF is temporally continuous and thus allows querying latent sharp images of any time within the exposure period $\mathcal{T}$. We formulate the IVF as a combination of the temporally constant component, \ie, the blurry image $\bf B$, and the temporally alternating component, \ie, $\phi_\theta$,
\begin{equation}\label{eq:ivf}
    {\bf I}(t) = {\bf B} + \phi_\theta (t;f_\gamma\left({\bf B}, \mathcal{E}_{\mathcal{T}}\right)),
\end{equation}
where $f_\gamma$ is a feature embedding function (with $\gamma$ denoting the parameters) to encode temporal variations in the feature domain in a higher dimensional space, and $\phi_\theta$ is a decoding function parameterized by an MLP (with $\theta$ as its parameters) to restore the temporal alternative component. Both $\phi_\theta$ and $f_\gamma$ are shared for any blurry image ${\bf B}$ and  events $\mathcal{E}_{\mathcal{T}}$.

\noindent\textbf{Relation to EDI~\cite{pan2019bringing}.} Events properly bridge the blurry image $\bf B$ and its latent images $\bf I$, providing a naive model for DeMo,
\begin{equation}\label{eq:edi}
\tilde{\bf I}(t) = \tilde{\bf B} - \tilde{\bf E}(t;\mathcal{E}_{\mathcal{T}}),
\end{equation}
with $\tilde{\bf B},\tilde{\bf I}$ and $\tilde{\bf E}$ are respectively the logarithms of $\bf B$, $\bf I$ and the event-based double integral (EDI) \cite{pan2019bringing}. The IVF in \cref{eq:ivf} and the EDI in \cref{eq:edi} can both tackle the DeMo task by shifting the original deblurring problem to the restoration of the temporal alternative component. However, EDI calculates the temporal alternative component by accumulating per-pixel events \cite{wang2020event}, and thus its performance might be degraded due to the event noise \cite{inivaiton2020} and incorrect estimation of the event threshold \cite{pan2020high}. Compared to EDI, our IVF attempts to learn the temporal alternative component by dual feature embedding (described in the next section), which leverages the merits of the blurry image $\bf B$ and events $\mathcal{E}_{\mathcal{T}}$ with the consideration of following issues: (1) suppression of event noise with the guidance of image features; (2) deblurring features extracted from the blurry input $\bf B$ with the enhancement of event features. 
% However, the event noise is severe both in spatial and temporal domains \cite{inivaiton2020}, which may be accumulated in EDI \cite{wang2020event}, resulting in considerable image degradation \cite{pan2020high}. %especially in the static background.

\noindent\textbf{Relation to INR~\cite{niemeyer2019,Bemana2020xfields}.} IVF learning is intuitively inspired by Implicit Neural Representation (INR), where videos can be represented in continuous form as a function of time and location \cite{niemeyer2019}. However, existing INR approaches generally require multi-frame inputs with clear information \cite{Bemana2020xfields}. Thus, learning an IVF from a blurry image is more challenging than existing INR learning tasks due to motion ambiguity and texture erasure. 

% The above two challenges are coupled together. On one hand, the high temporal resolution of events provides the intra-frame information that can effectively boost the performance of DeMo. On the other hand, the learned DeMo implicitly reflects scene dynamics which can be employed to reduce event noise. Thus, one should complete the task of DeMo with attention to both event noise suppression and blurry feature enhancement.

%Thus,  (3) jointly estimating the TA component with the fusion of enhanced event features and deblurred image features. 
% IVF \cref{eq:ivf} and EDI \cref{eq:edi} can tackle the DeMo task by shifting the original deblurring problem to the restoration of the TA components. The difference is that the IVF attempts to learn the TA components by dual feature embedding, which leverages the merits of the blurry image $\bf B$ and the concurrent events $\mathcal{E}_{\mathcal{T}}$ for (1) suppressing event noises with the guidance of image features; (2) deblurring  features extracted from $\bf B$ with the enhancement of event features; (3) jointly estimating the TA components with the fusion of event features and deblurred image features. However, EDI calculates the TA components only by accumulating events in a pixel-wised manner. Thus to some extent, the IVF can be considered as a generalization of the EDI model in the intensity domain (EDI is in the logarithmic domain), but more robust to event noises than EDI since it combines information from the blurry image and events.
\cref{fig:overview_dals} illustrates the proposed network to fulfill the IVF in \cref{eq:ivf}. Accordingly, the architecture of IVF contains two modules, \ie, the module of dual feature embedding $f_\gamma$ and the module of continuous-time decoding MLP $\phi_\theta$.

\subsection{Dual Feature Embedding Network}
\label{sec:DFEN}
% As discussed in \cref{sec:ivf}, we are interested in developing a network to take the merits of both event streams and images for feature embedding. Accordingly, we build our Dual Feature Embedding Network (DFEN) to embed those two sources of information in a unified embedding space.
To achieve the mutual compensation of events and frames, we design a Dual Feature Embedding Network, \ie, DFEN, to embed those two sources of information in a unified embedding space.
%添加event representation的说明
Our DFEN initializes the input blurry image $\mathbf{B}$ and the event tensor $\mathbf{E}$ by two separate Shallow Feature Extraction (SFE) modules using the pixel-unshuffle layer~\cite{shi2016} with a stride of $2$ followed by two convolution blocks to extract the shallow features: %of image and the events into feature maps with $C$ channels, denoted by%
\begin{equation}
    \begin{aligned}
    \mathbf{B/E}^{\downarrow} &= \text{Pixel-Unshuffle}( \mathbf{B/E} ), \\
    \mathbf{B/E}_{feat}^{-1} &= \text{Conv}_{5\times5}( \mathbf{B/E}^{\downarrow} ), \\
    \mathbf{B/E}_{feat}^{0} &= \text{Conv}_{3\times3}( \mathbf{B/E}_{feat}^{-1} ), \\   
    \end{aligned}
\end{equation}
where $\mathbf{B/E}^{\downarrow}$ denotes the downsampled blur/event frame and $\mathbf{B/E}_{feat}^{-1}, \mathbf{B/E}_{feat}^{0} \in \mathbb{R}^{C\times \frac{H}{2} \times \frac{W}{2}}$ denote the feature maps of blurry image/events with $C$ channels.

\begin{figure}[t]
\centering
\includegraphics[width=0.45\textwidth]{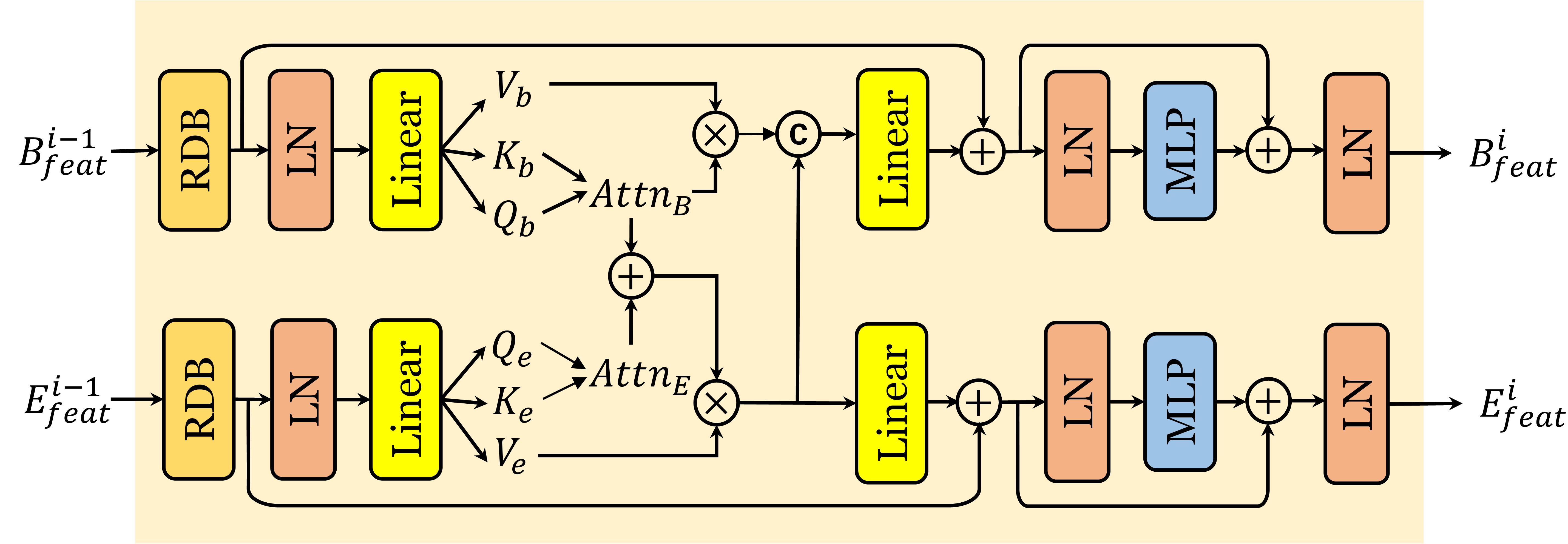}
\caption{Detailed Dual Attention block to take into account the Latent Structures (DALS) shared with the dual input features $\mathbf{B}_{feat}$ and $\mathbf{E}_{feat}$.}
\label{fig:dals}
\end{figure}

% With the initialized features $\mathbf{B}_{feat}, \mathbf{E}_{feat}$,
We then leverage several Dual Attention blocks to take into account the Latent Structures (DALS) shared with the dual input features, as shown in \cref{fig:dals}. Each DALS block is composed of a
{\color{black}{Residual Dense Block (RDB)}} \cite{zhang2018residual} and a Window-based Multi-head Self-Attention (W-MSA) block in a sequential manner, where a Dual Attention Mechanism (DAM) is implemented in DALS for event noise suppression and blurry feature enhancement.

% we use the Window-based Multi-Head Self Attention (W-MSA) layers to exploit the latent structures from two information sources, which finally yields a unified feature embedding for the dual inputs.

% With the initialized features $\mathbf{B}_{feat}, \mathbf{E}_{feat}$, we use the Window-based Multi-Head Self Attention (W-MSA) layers to exploit the latent structures from two information sources, which finally yields a unified feature embedding for the dual inputs.

\noindent \textbf{Window-based Multi-head Self-Attention (W-MSA).} We follow the basic strategy proposed in \cite{liu2021c} to first partition the features $\mathbf{B/E}_{feat}^{0}$ into $L$ local patches, and then calculate the self-attention by $$\text{W-MSA}(\text{Attn}, V) = \text{Attn} \cdot V,$$ with the self-attention weight defined as $\text{Attn}=\text{softmax}(\frac{QK^T}{\sqrt{d_k}})$, where $Q$, $K$ and $V$ are the encoded queries, keys, and values, yielded by an MLP for each of them. Different from the case of uni-modal input in \cite{liu2021c}, we further propose the DAM to modify the self-attention weights for mutual compensation of multi-modal signals.

% For a feature map $\mathbf{F}\in\mathbb{R}^{C\times h\times w}$, we follow \cite{liu2021c} to partition the features $\mathbf{F}$ into $L$ local patches, denoted by $\mathbf{F}_{0},\ldots, \mathbf{F}_{L-1} \in \mathbb{R}^{C\times m\times m}$. 
% For each feature patch $\mathbf{F}_i$, we first compute the self-attention weight by $\text{Attn}=\text{softmax}(\frac{QK^T}{\sqrt{d_k}})$, and then generate the self-attention by 
% $
%     \text{W-MSA}(\text{Attn}, V) = \text{Attn} \cdot V,
% $
% where $Q$, $K$ and $V$ are the encoded queries, keys and values, yielded by MLP for each feature.

% \noindent \textbf{Window-based Multi-head Self Attention.} For a feature map $\mathbf{F}\in\mathbb{R}^{C\times h\times w}$, we follow \cite{liu2021c} to partition the features $\mathbf{F}$ into $L$ local patches, denoted by $\mathbf{F}_{0},\ldots, \mathbf{F}_{L-1} \in \mathbb{R}^{C\times m\times m}$. For each feature patch $\mathbf{F}_i$, we compute the attention by
% $
%     \text{W-MSA}(Q, K, V) = \text{softmax}(\frac{QK^T}{\sqrt{d_k}}) V,
% $
% where $Q$, $K$ and $V$ are the encoded queries, keys and values, yielded by MLP for each feature.

% \noindent \textbf{Dual Attention with Latent Structures.} 
% As shown in \cref{fig:overview_dals}, we leverage several Dual Attention blocks to take into account Latent Structures (DALS) shared with the dual input features $\mathbf{B}_{feat}$ and $\mathbf{E}_{feat}$. Each DALS block is composed of a
% {\color{black}{Residual Dense Block (RDB)}} \cite{zhang2018residual} followed by a W-MSA block in a sequential manner. 
\noindent \textbf{Dual Attention Mechanism.} 
In the $i$-th ($1\leq i \leq N$) W-MSA block, two parallel paths are designed to first compute the self-attention weights $\text{Attn}_{E}$ and $\text{Attn}_{B}$ for the features of the events and the image. We then propose a dual attention mechanism to mutually compensate for the event features and image features.
Firstly, as the input events usually contain unexpected noise, we use the learned attention weight $\text{Attn}_{B}$ from the blurry features (less noisy) to calibrate the attention weights $\text{Attn}_{E}$ by 
\begin{equation}\label{eq:atten_e0}
    \mbox{Attn}_{E} \leftarrow \mbox{Attn}_{E} + \mbox{Attn}_{B}.
\end{equation}
Once the attention weight $\text{Attn}_{E}$ is calibrated, we compute the event features from $\mathbf{E}_{feat}^{i-1}$ by
\begin{equation}\label{eq:atten_e}
\begin{split}
\bar{\mathbf{E}}_{feat}^{i-1} &= \text{W-MSA}(\mbox{Attn}_E^{i-1},V(\mathbf{E}_{feat}^{i-1})),
% \mbox{Attn}_E^{i-1} \cdot V(\mathbf{E}_{feat}^{i-1}), 
\\
\mathbf{E}_{feat}^i &= \text{MLP}_E^{i-1}(\bar{\mathbf{E}}_{feat}^{i-1}),
\end{split}
\end{equation}
where $V(\cdot)$ is the corresponding value operator. 
Such a design suppresses the incorrectly-estimated attention for the noisy events, leading to the enhancement of event features from the blurry image. %({\bf B2E}).
On the other hand, the image features often suffer from the loss of textures due to the blur degradation, which can be potentially compensated by events.
% extracted from the blurry inputs are often  suffer from blurry textures, which can be further eliminated benefiting from the low latency of events. 
To achieve this end, we compute the image features from both $\mathbf{B}_{feat}^{i-1}$ and $\bar{\mathbf{E}}_{feat}^{i-1}$ by
\begin{equation}\label{eq:dual-attn}
\begin{split}
    \bar{\mathbf{B}}_{feat}^{i-1} &=
    \text{W-MSA}(\text{Attn}_B^{i-1},V(\mathbf{B}_{feat}^{i-1})),
    % \text{Attn}_B^{i-1} \cdot V(\mathbf{B}_{feat}^{i-1}), 
    \\
    \mathbf{B}_{feat}^i &=\text{MLP}_B^{i-1}([\bar{\mathbf{B}}_{feat}^{i-1};\bar{\mathbf{E}}_{feat}^{i-1}]),
\end{split}
\end{equation}
where $\bar{\mathbf{E}}_{feat}^{i-1}$ is the weighted event features computed from \cref{eq:atten_e}. The feature concatenation compensates the image features with the texture information from the event features, leading to deblurring effects. %({\bf E2B}).
% the texture ambiguities, leading to deblurring effects from event features ({\bf E2B}).

% Then 
% \begin{equation}
%     \begin{split}
%     \mathbf{B}_{feat}^i &= \text{MLP}_B^{i-1}([\mathbf{B}_{feat}^{i-1};\mathbf{E}_{feat}^{i-1}])\\
%     \mathbf{E}_{feat}^i &= \text{MLP}_E^{i-1}(\mathbf{E}_{feat}^{i-1})
% \end{split},
% \end{equation}

Combining Eqs.~\eqref{eq:atten_e0}, \eqref{eq:atten_e}, and \eqref{eq:dual-attn}, we firstly calibrate the attention weights $\text{Attn}_E$ for events by using the contextual information from the blurry image features to obtain better event features.
 Then, the enhanced event features are concatenated with the initial image features for further refinement. As the partitioning operation splits an image plane into non-overlapped regions, we use the shifting operation for the blocks with even indices to keep the consistency between non-overlapped patches. 
%  In summary, we call this mechanism as Dual Attention with Latent Structures (DALS). 
 Finally, we concatenate the output image features (with the inverse partitioning operation) of all DALS blocks into $\mathbf{F}_{cat} \in \mathbb{R}^{NC\times \frac{h}{2} \times \frac{w}{2}}$ as the feature embedding. In our implementation, the number of DALS blocks is set to $20$. 
% Nan: do we need to use ablation study to justify our design?
    
% \paragraph{Event Noise Suppression via Cross-Modality Attention}
% 通过image的attention weight，加权event，去噪声

% \paragraph{Feature Deblurring via Post-Attention Fusion}
% 通过event attention之后的结果，对image的feature（模糊）进行多级融合去模糊
% 这里我说的是前面每个W-MSA的V的融合
With the learned high-dimensional feature embedding, we combine the initial features $\mathbf{B}_{feat}^{-1}$ extracted from the blurry image with the outputs of all W-MSA blocks {\color{black}{by a Global Feature Fusion (GFF) module}}. Specifically, we use a $1\times 1$ convolution layer to firstly reduce the feature channels from $N\times C$ to $C$ and then use a $k\times k$ convolution layer to generate the deblurred features $\mathbf{F}_{db}$ by
\begin{equation}
    \mathbf{F}_{db} = \mathbf{B}_{feat}^{-1} + \text{Conv}_{k\times k}(\text{Conv}_{1\times 1}(\mathbf{F}_{cat})).
\end{equation}

\subsection{Continuous-Time Decoding MLP}

\begin{figure}[t]
\centering
\includegraphics[width=0.35\textwidth]{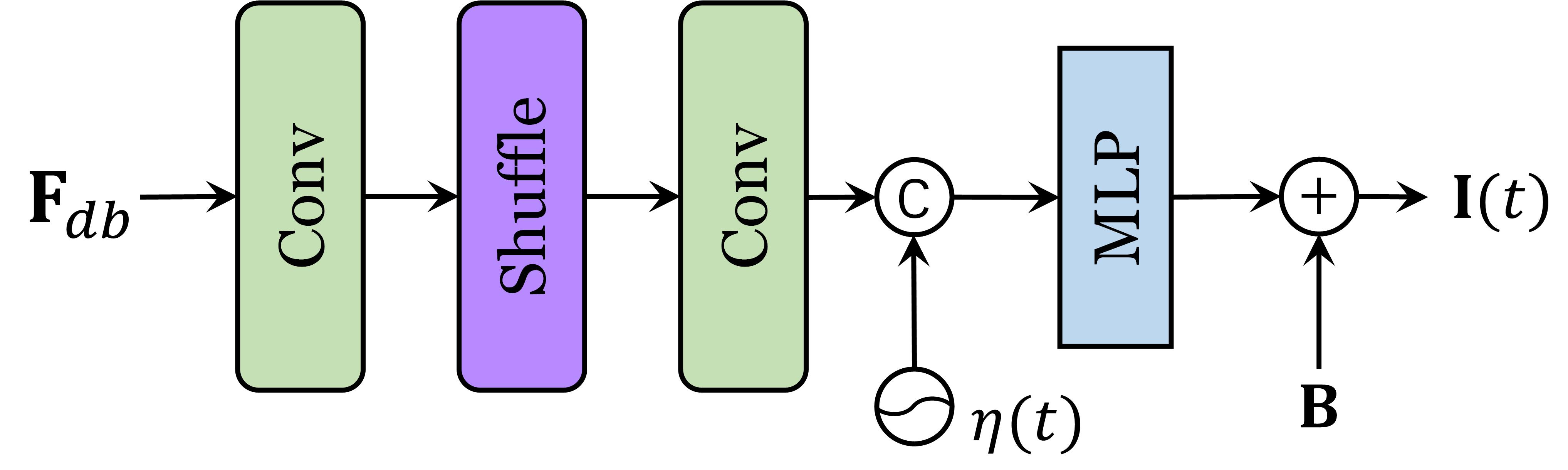}
\caption{Details of the continuous-time decoding MLP, which accepts the deblurred feature $\mathbf{F}_{db}$ output by DFEN and the embedding of any specified time $\eta(t)$ to decode the latent sharp image $\mathbf{I}(t)$.
} 
\label{fig:continuous-time decoding}
\end{figure}
As our dual feature embedding network encodes the temporal information, we propose to query the sharp image at an arbitrary timestamp in the range of the imaging exposure period. Without loss of generality, we normalize the exposure time interval into $[0, 1]$ and query the normalized timestamp $t \in [0,1]$. For the query time $t$, we encode it into a high-dimensional vector $\eta(t) \in \mathbb{R}^{2L}$ ($L=8$ in our experiment) following the Fourier encoding scheme used in \cite{benbarka2021},
\begin{equation}\small 
    \eta(t) = 
    \begin{pmatrix}
    \cos(2^0\pi t), \sin(2^0\pi t), \ldots, \cos(2^{L-1}\pi t), \sin(2^{L-1}\pi t)
    \end{pmatrix}.
\end{equation}
After decoding the query image at timestamp $t$, we apply a pixel-shuffle layer and convolution layers to upsample the deblurred feature $\mathbf{F}_{db}$ into $\mathbf{F}_{db}^{\uparrow} \in \mathbb{R}^{32\times H\times W}$ to keep the original image resolution. Following that, we concatenate the time embedding $\eta(t)$ in each pixel to yield a time-specific feature map and then use an MLP with four $256$-D hidden linear layers to finally decode the sharp latent frame at time instance $t$, denoted by $\mathbf{I}(t)$. Unlike the previous approaches that only produce the latent images at fixed timestamps, our time-continuous decoding module is able to restore the sharp images at any given time instances $t$.

 Given a single blurry image $\mathbf{B}$ and its concurrent event stream $\mathcal{E}_\mathcal{T}$, one can first use the DFEN module for feature embedding and then get the deblurred features $f_\gamma(\mathbf{B},\mathcal{E}_\mathcal{T})=\mathbf{F}_{dblr}$, which are time constant. In the continuous-time decoding MLP, we only need to change the value of time $t$ to restore the temporally alternating component $\phi_\theta$, and then add it to the blurry image $\mathbf{B}$ to restore the latent image $\mathbf{I}(t)$. 
 % Thus, once the feature embedding is learned, the inference of $\mathbf{I}(t)$ by DeMo is computationally efficient. \textcolor{red}{(ZX: should we add a computational complexity exp to support this claim?)}

%\input{fig/gopro}
% In our setting, $L$ is equal to $3$ and $\left [ P_1, P_2, P_3 \right ]$ is equal to $\left [ 3000, 5000, 7000\right ]$.
% and that the events accumulated over a short period of time exactly reflect the appearances of the image edges at the corresponding moments.

% supervision by finite frames (GT)

% flow consistency

% intensity consistency
% \begin{table}
% \centering
% \resizebox{\linewidth}{!}{ %< auto-adjusts font size to fill line
% \begin{tabular}{@{}lccc@{}}
% \toprule
% Method & Frobnability & Frobnability & Frobnability \\
% \midrule
% Theirs & Frumpy \\
% Yours & Frobbly \\
% Ours & Makes one's heart Frob\\
% \bottomrule
% \end{tabular}
% } % \resizebox
% \caption{
% % 
% \textbf{State of the art} -- Glory to the bold numbers.
% % 
% } % \caption
% \label{tab:sota}
% \end{table}

% \usepackage{multirow}
% \usepackage{multirow}
% \usepackage{multirow}
\begin{table*}[!thb]
    \centering
    \caption{Quantitative comparisons of DeMo-IVF to the state-of-the-art motion deblurring methods on restoring a single image and a video sequence (\ie, 7 images). For the single image restoration, all methods are evaluated with respect to the middle frame of the sequence prediction except for eSL-Net which recovers the first frame according to \cite{wang2020event}. The methods eSL-Net, LEDVDI, and RED-Net are all fine-tuned on the training sets of the REDS and HQF datasets.
    % Note that the $*$ mark denotes that the corresponding methods are fine-tuned on the training set of GoPro and HQF datasets.
    % methods with the $*$ mark are fine-tuned on the training set of GoPro and HQF datasets.
    }
    \renewcommand\arraystretch{1.2}
    \label{tab:sota}
    \begin{tabular}{lccccccccccc}
        \hline
        \multirow{3}{*}{Method} & \multicolumn{5}{c}{Single frame restoration} & & \multicolumn{5}{c}{7 frames restoration} \\ 
        
        \cline{2-6} \cline{8-12}
        & \multicolumn{2}{c}{REDS} & & \multicolumn{2}{c}{HQF} & & \multicolumn{2}{c}{REDS} &  & \multicolumn{2}{c}{HQF} \\ 
        
        \cline{2-3} \cline{5-6} \cline{8-9} \cline{11-12}
        & PSNR$\uparrow$ & SSIM$\uparrow$ & & PSNR$\uparrow$ & SSIM$\uparrow$ & & PSNR$\uparrow$ & SSIM$\uparrow$ & & PSNR$\uparrow$ & SSIM$\uparrow$ \\
        
        \hline
        LEVS \cite{jin2018learning}     & 21.885 & 0.6243 &   & 21.900 & 0.6367 &   & 19.851 & 0.5288 &   & 19.068 & 0.5403 \\
        Motion-ETR \cite{zhang2021exposure}    & 22.305 & 0.6494 &   & 22.516 & 0.6450 &   & 19.543 & 0.5064 &   & 18.930 & 0.5239 \\
        EDI \cite{pan2019bringing}      & 21.517 & 0.6409 &   & 20.321 & 0.6212 &   & 20.939 & 0.6176 &   & 19.081 & 0.5873 \\
        eSL-Net \cite{wang2020event}    & 24.791 & 0.8009 &   & 20.438 & 0.6017 &   & 23.955 & 0.7578 &   & 19.866 & 0.5851 \\
        E-CIR \cite{song2022cir}        & 26.541 & 0.7898 &   & 25.851 & 0.7819 &   & 26.287 & 0.7734 &   & 25.221 & 0.7525 \\
        LEDVDI \cite{lin2020learning}   & 27.818 & 0.8190 &   & 27.656 & 0.8325 &   & 27.884 & 0.8251 &   & 28.208 & 0.8413 \\
        RED-Net \cite{xu2021motion}     & 29.955 & 0.8704 &   & 29.543 & 0.8646 &   & 29.431 & 0.8619 &   & 28.667 & 0.8544 \\
        % LEDVDI \cite{lin2020learning}   & 22.856 & 0.7334 &   & 22.221 & 0.7567 &   & 22.673 & 0.7329 &   & 21.558 & 0.7355 \\
        % RED-Net \cite{xu2021motion}     & 28.984 & 0.8499 &   & 25.717 & 0.7629 &   & 28.343 & 0.8359 &   & 24.076 & 0.7340 \\
        % LEDVDI \cite{lin2020learning} &  24.325      & 0.8286      &   & 25.398     & 0.8456     &  &  24.038      & 0.8248      &   &  24.302     & 0.8318     \\
        % RED-Net \cite{xu2021motion}     & 28.984      &  0.8499      &   &  26.094     & 0.8495     &  & 28.343      & 0.8359      &   & 24.368     & 0.8214     \\ 
        \hline 
        Ours & \bf{31.584} & \bf{0.9075} &   & \bf{30.877} & \bf{0.8914} &   & \bf{31.111} & \bf{0.8995} &   & {\bf 30.117} & {\bf 0.8843} \\ 
        % DeMo-IVF & \bf{30.701} & \bf{0.9364} &   & - & - &   & \bf{30.182} & \bf{0.9303} &   & - & - \\ 
        \hline
    \end{tabular}
\end{table*}

\subsection{Motion and Texture Guided Supervisions}
% We learn the implicit video function with sparsely sampled sharp images captured in $K$ timestamps $T_K = \{t_i\}_{i=1}^{K}$ in the exposure interval.  Let $\mathcal{\hat{I}} = (\mathbf{\hat{I}}_{t_1}, \ldots \mathbf{\hat{I}}_{t_K})$ be the ground-truth images. It is straightforward to use the $\ell_1$ loss to optimize our DFEN $f_\gamma$ and decoding MLP $\phi_\theta$, \ie,
% \begin{equation}\label{eq:loss-im}
%     \mathcal{L}_{im} = \frac{1}{K} \sum_{i=1}^K \|{\bf I}(t_i) - \mathbf{\hat{I}}_{t_i}\|_1.
% \end{equation}

% For a single blurry image $\mathbf{B}$ and the corresponding event stream $\mathcal{E_T}$, there exists $K$ timestamps $T_K = \{t_i\}_{i=1}^{K}$ and ground-truth images $\mathcal{\hat{I}} = (\mathbf{\hat{I}}_{t_1}, \ldots \mathbf{\hat{I}}_{t_K})$ in the exposure interval.
%pairs of timestamps and ground-truth images $\left\{ \left( t_k, \mathbf{\hat{I}}(t_k) \right) | 1 \le k \le K, t_k \in \mathcal{T} \right\}$. 

% Practically, we only have ground-truth reference images of limited timestamps during the exposure time. Therefore, 

Denoting $\mathcal{\hat{I}} = (\mathbf{\hat{I}}({t_1}), \ldots ,\mathbf{\hat{I}}({t_K}))$ as the ground-truth images of $K$ referenced timestamps $\mathcal{T}_K \triangleq \{t_k\}_{k=1}^{K}$, we can train the Implicit Video Function (IVF) composed of the DFEN $f_\gamma$ and the decoding MLP $\phi_\theta$, using the $\ell_1$ loss,

% We learn the implicit video function with sampled sharp images captured in $K$ timestamps $T_K = \{t_i\}_{i=1}^{K}$ in the exposure interval. Let $\mathcal{\hat{I}} = (\mathbf{\hat{I}}({t_1}), \ldots \mathbf{\hat{I}}({t_K}))$ be the ground-truth images.
% It is straightforward to use the $\ell_1$ loss to optimize our DFEN $f_\gamma$ and decoding MLP $\phi_\theta$, 
\begin{equation}\label{eq:loss-im}
    \mathcal{L}_{im} = \frac{1}{K} \sum_{k=1}^K \|{\bf I}(t_k) - \mathbf{\hat{I}}({t_k})\|_1.
\end{equation}
It is straightforward that the trained IVF model can achieve better performance with more supervision, provided by ground-truth images of more referenced timestamps. However, the referenced timestamps would be limited due to the practical frame-rate constraint. On the other hand, only the supervision of referenced timestamps would lead to imbalanced performance between restored latent images of the referenced and the non-referenced timestamps. In this subsection, we will address this problem by simultaneously employing motion- and texture-
guided supervisions.
% as $K$ increases, \ie, more data with ground-truth images of different timestamps are used for supervision, the trained IVF model will achieve the continuous-time video extraction task better.
% However, due to practical constraints such as limited camera frame rates, for one single blurry image, we can only acquire a few pairs of timestamps and sharp images as the supervision signals.  
% Restricted by the sparse sampling, only using \cref{eq:loss-im} over all the supervision samples is not enough for our goal of continuously decoding sharp images at any given time $t$. And in our experiments, we found that such a treatment will lead to a bias in the sparsely sampled timestamps. 
% Therefore, the continuity of the time should be considered in the training phase of the network. 

% \vspace{-1em}
% \paragraph{Flow Consistency} 
% We address the bias issue caused by the sparse samples using event-based optical flow. 
% Ideally, if an optical flow is obtained from the already-sampled time $t_k$ to any time $t = t_k + \delta$, it is feasible to warp the sharp image $\mathbf{\hat{I}}_{t_k}$ to a sharp image at time $t$ as the  ground-truth for learning. 
% However, due to the lack of optical flow between any timestamp, we alternatively use an off-the-shelf method EV-Flow~\cite{zhu2018ev} to yield the pseudo flow maps $\text{Flow}_{t_i\to t}=\text{EV-Flow}(\mathcal{E}_{t_i\to t})$, with $\mathcal{E}_{t_i\to t}$ denoting events triggered during $[t_i, t]$. %\textcolor{red}{Nan: please add the detail of how to sample time $t$ for flow consistency.}
\noindent{\bf Motion-Guided Supervision.}
The motion-guided supervision allows the model to exploit motion continuity to achieve continuous-time video extraction. The key idea is utilizing the motion information encoded in events to bridge the inter-frame connection between latent sharp images.
Given an optical flow $\text{Flow}_{t_i \to t_i + \delta}$ obtained from the referenced timestamp $t_i$ to any non-referenced timestamp $t = t_i + \delta$, it is feasible to warp $\mathbf{\hat{I}}(t_i)$ to the latent restoration of the non-referenced time $t$,
\begin{equation}\label{eq:motion-guided}
    \mathbf{\hat{I}}(t_i + \delta) = \text{Warp} \left ( \mathbf{\hat{I}}(t_i), \text{Flow}_{t_i \to t_i + \delta} \right ).
\end{equation}
By establishing the above motion connections, the latent restorations of the non-referenced timestamps can be supervised.
To achieve the end, we directly employ the off-the-shelf method EV-Flow~\cite{zhu2018ev} to yield the flow maps $\text{Flow}_{t_i\to t}=\text{EV-Flow}(\mathcal{E}_{t_i\to t})$, with $\mathcal{E}_{t_i\to t}$ denoting events triggered during $[t_i, t]$. %\textcolor{red}{Nan: please add the detail of how to sample time $t$ for flow consistency.}
% In the training phase, we randomly select $M$ non-referenced timestamps $\{t_j^{\prime} \}_{j=1}^{M}$ and compute the motion loss to calibrate the corresponding latent restorations,
% \begin{equation}  
%     \mathcal{L}_{motion} = \frac{1}{M \times N} \sum_{j=1}^M \sum_{k=1}^N \left\|{\bf I}(t_j^\prime) - \hat{{\bf I}}^k({t_j^\prime})\right\|_1.
% \end{equation}

% In detail, we randomly sample $M$ timestamps $\{t_j^{\prime} \}_{j=1}^{M}$ 
% (For clarity, we refer to the timestamps in $T_k$ where real ground truth can be obtained as the Referenced timestamps and the timestamps outside of it as the Non-referenced timestamps.).
For each non-referenced timestamp $t_j^{\prime} \notin \mathcal{T}_K$, we select $N$ nearest referenced timestamps $ \{t_{j, k} \}_{k=1}^{N}$ from $\mathcal{T}_K$, \ie, $t_{j, k} \in \mathcal{T}_K$, where $N \leq K$. The corresponding optical flow from $t_{j,k}$ to $t_j^{\prime}$ is calculated with the in-between events, \ie, $\mathcal{E}_{t_{j,k}\to t_j^{\prime}}$. Then we can warp the ground-truth images $\mathbf{\hat{I}}({t_{j,k}})$ of the referenced timestamps $t_{j,k}$ to obtain the supervision $\mathbf{\hat{I}}^k({t_j^\prime})$ of the non-referenced timestamp $t_j^{\prime}$ according to \cref{eq:motion-guided}. In the training phase, we randomly select $M$ non-referenced timestamps $\{t_j^{\prime} \}_{j=1}^{M}$ and thus compute the motion-guided loss as the following,
% We use events to compute the optical flow from $t_{j,k}$ to $t_j^{\prime}$ and warp the image $\mathbf{\hat{I}}({t_{j,k}})$ to $\mathbf{\hat{I}}^k({t_j^\prime})$, and then compute the motion loss to calibrate the image prediction without references,
\begin{equation}  
    \mathcal{L}_{motion} = \frac{1}{M \times N} \sum_{j=1}^M \sum_{k=1}^N \left\|{\bf I}(t_j^\prime) - \hat{{\bf I}}^k({t_j^\prime})\right\|_1.
\end{equation}
% where $\hat{{\bf I}}^k({t_j^\prime}) = \text{Warp}(\mathbf{\hat{I}}({t_{j, k}}), \text{Flow}_{t_{j, k}\to t_j^\prime})$.
Jointly supervising ${\bf I}(t_j^\prime)$ by multiple $\{ \hat{{\bf I}}^k({t_j^\prime}) \}_{k=1}^{N}$ warped from $N$ ground-truth images at different referenced timestamps can help to alleviate distortions caused by optical flow errors.

\noindent \textbf{Texture-Guided Supervision.} 
The performance of both continuous-time video restorations and motion predictions largely relies on the extremely high temporal resolution of events. However, we cannot fully utilize this property when feeding events into our proposed IVF and the motion prediction network, \ie, EV-Flow, since events should be stacked into tensors to match the CNN inputs. The temporal information might be lost when stacking events, leading to artifacts and blurry restorations, especially in regions with abundant textures and large motions. Thus the texture-guided supervision is employed further to enhance the overall sharpness of the continuous-time video restorations. 
% On the one hand, the motion-guided loss is largely dependent on the accuracy of the predicted optical flow. 
% To alleviate the impact of these two factors, we further propose texture-guided refinement to compensate for the loss of intermediate information and to recover better textures and details.
In detail, we introduce an Event-based Edge Refinement ($\operatorname{EER}$) module to refine the initial deblurring results ${\mathbf I}(t)$ with the guidance of the events, \ie, 
\begin{equation}\label{eq:eer}
    {\mathbf I}_{\mathrm{refine}}(t) = \operatorname{EER} \left ( {\mathbf I}(t),  \mathcal{E}_t \right ),
\end{equation}
where ${\mathbf I}_{\mathrm{refine}}(t)$ indicates the refined image result and $\mathcal{E}_t$ indicates the subtle event segment to refine the restoration of the timestamp $t\in \mathcal{T}$.
% 定义local event
Specifically, we define $\mathcal{E}_{\ge t}^{P}\subset \mathcal{E}_{\mathcal{T}}$ ($\mathcal{E}_{\le t}^{P}\subset \mathcal{E}_{\mathcal{T}}$) the set of temporally nearest $P$ events with timestamps greater (smaller) than $t$, \ie, 
\[
\begin{split}
   \mathcal{E}_{\ge t}^{P} &\triangleq \left\{ ({\bf x}_i,p_i,t_i) | t_i\in [t,t_P], i\in \{1,2,...,P\}\right\}, \\
   \mathcal{E}_{\le t}^{P} &\triangleq \left\{ ({\bf x}_i,p_i,t_i) | t_i\in [t_P,t), i\in \{1,2,...,P\}\right\}, \\
\end{split}
\]
with $t_P$ timestamp of the $P$-th temporally nearest event. Then the subtle event segment $\mathcal{E}_t$ is defined as follows, 
\begin{equation}
    \mathcal{E}_t = \left \{ \mathcal{E}_{\ge t}^{P_1}, \mathcal{E}_{\le t}^{P_1}, ..., \mathcal{E}_{\ge t}^{P_L},  \mathcal{E}_{\le t}^{P_L}  \right \},
\end{equation}
where $2L$ subsets with different numbers of events are grouped together to provide rich and accurate information at different temporal scales. Note that $\mathcal{E}_t$ is only parameterized by the timestamp $t$ without polarity reversal operations, thus avoiding the modeling error of the event re-shuffle process utilized in~\cite{wang2020event,zhang2022unifying}.
The $\operatorname{EER}$ module \cref{eq:eer} is implemented based on the RDN~\cite{zhang2018residual} backbone in our setting and fed with input by concatenating the initial deblurring result ${\mathbf I}(t)$ and the corresponding subtle event segments $\mathcal{E}_t$.
% extracts the edge structures from the events to refine the initial deblurring results. 
% We implement it as the neural network and learn the task from data.
% Unlike the training process of the IVF model, the data on the non-key-timestamps is not used here, in order to avoid the introduction of edge blur by interpolation.

% 与已有方法进行比较
% Our EER module shares the same key idea as the DEF module proposed in LEMD\cite{jiang2020learning}, which enhances the reconstruction with the aid of sharp edge prior extracted from events. 
% DEF gets the edge guidance map by sampling events between two latent sharp images, while EER directly uses events near the enhanced target to provide edge priors, which is easier to implement and avoids sampling errors.
% Moreover, DEF is incorporated into the main network to provide edge priors for the reconstruction of each predefined timestamp, which is not feasible for IVF because the target of the reconstruction is determined by the additional input $t$.
% Therefore, the EER is designed as a stand-alone module that refines the initial estimates, which is more compatible with the IVF model and can be cascaded to any other event-based deblurring method for edge refinement.

\noindent \textbf{Training Strategy.}
The whole training process is composed of two phases. 
In the first phase, we optimize our IVF model using $\mathcal{L}_{im}$ and $\mathcal{L}_{motion}$, 
% Stage 1: 
% The final loss function is 
% % \input{tab/sota.tex}
\begin{equation}\label{eq:allloss}
% \small 
    \mathcal{L}_{total} =\lambda_1 \mathcal{L}_{im}  + \lambda_2 \mathcal{L}_{motion},
\end{equation}
with $\lambda_1$ and $\lambda_2$ being the balancing parameters.
The model trained in the first stage is able to predict the continuous illumination change for each pixel from the blurry image with events, thus restoring the latent sharp images at arbitrarily specified timestamps. 
In the second phase, we fix the parameters of the well-optimized IVF model and use it to estimate the initial deblurring results of the reference timestamps $t_k\in \mathcal{T}_K$, which are then fed into the $\operatorname{EER}$ module along with the corresponding events $\mathcal{E}_{t_k}$.
\begin{equation}
    {\mathbf I}_{\mathrm{refine}}(t_k) = \operatorname{EER} \left ( \mathbf{I}(t_k),  \mathcal{E}_{t_k} \right ),
\end{equation}
where $\mathbf{I}(t_k)$ is learned IVF of the continuous-time video at timestamp $t_k$ according to \cref{eq:ivf}.
The $\ell_1$ loss between the refined results ${\mathbf I}_{\mathrm{refine}}(t_k)$ and the ground truth images $\mathbf{\hat{I}}({t_k})$ is used to optimize the $\operatorname{EER}$ module,
\begin{equation}
    \mathcal{L}_{texture} = \frac{1}{K} \sum_{k=1}^K \|{\mathbf I}_{\mathrm{refine}}(t_k) - \mathbf{\hat{I}}({t_k})\|_1.
\end{equation}
Note that the performance gap between reconstructions of reference and non-reference timestamps has been narrowed by motion-guided supervision $\mathcal{L}_{motion}$ in the first stage of training. Thus, although only the images at the reference timestamps are used in the second stage, our texture-guided supervision can lead to a general improvement in the restored images at both reference and non-reference timestamps.
% Note that with the introduction of motion-guided supervision in the first stage of training, the performance gap between reconstructions of reference and non-reference timestamps have been greatly alleviated.
% Although only the data of reference timestamps is used for supervision in the second stage, our EER module can generally refine the sharp images at non-reference timestamps.

After the two stages of training, the overall network
% IVF and EER together 
achieves extracting continuous-time sharp video with delicate details from the blurry image and events.

% 说明refine只用了reference监督，但是可以对non-reference起作用的原因

\input{fig/fig_reds_hqf.tex}
\input{fig/fig_reds_31.tex}
\input{fig/fig_rbe.tex}

%%%%%%%%%%%%
% Results  %
%%%%%%%%%%%%

\section{Experiments and Analysis}
% \vspace{-1cm}
% \input{fig/fig2.tex}

This section evaluates and analyzes the proposed DeMo-IVF method. In \cref{sec:6-1}, we first present the experimental settings, including the datasets and implementation details. The performance of state-of-the-art methods and our proposed DeMo-IVF are then compared in \cref{sec:6-2}, on restoring single frame, video sequence, and continuous-time videos. 
After that, we analyze the effectiveness of network architecture and training strategy of our proposed DeMo-IVF method respectively in~\cref{sec:6-3}.

\subsection{Experimental Settings}\label{sec:6-1}
\subsubsection{Datasets} 
% ZX =====================
Three different datasets are employed to evaluate the proposed DeMo-IVF, including the synthetic {REDS} dataset with synthesized blurry images and events based on the REDS dataset~\cite{nah2019ntire}, the semi-synthetic {HQF}~\cite{stoffregen2020reducing} with synthesized blurry images and real-world events captured with a DAVIS346 event camera, and the real-world {RBE} dataset~\cite{xu2021motion} with real-world blurry images and events captured with a DAVIS346 event camera. 

\noindent{\bf REDS.} The original REDS dataset \cite{nah2019ntire} contains 270 videos captured at 120 fps and each video contains 500 sharp and clear images at a resolution of $720 \times 1280$. To imitate the output of real event cameras, we first convert all videos to grayscale image sequences, downsample them to $180 \times 320$, and increase the video frame rate to 480 fps with the leading video frame interpolation algorithm~\cite{huang2022real}. 
After that, we simulate both events and blurry images based on the high frame rate video sequences, where
% we use the video frame interpolation algorithm~\cite{huang2022real} to obtain high frame-rate videos with 480 fps, which are applied for event generation and blurry synthesis. 
the ESIM~\cite{rebecq2018esim} simulator is adopted to generate event streams and 121 consecutive frames are averaged to synthesize blurry images. Thus each blurry image corresponds to 31 sharp frames in the original captured videos and we define them as the ground truths. We follow \cite{nah2019ntire} to split the REDS dataset into the training and testing sets respectively with 240 and 30 videos. 
% To facilitate algorithm comparisons on continuous-time video restoration (\cref{comparison}), we use 31 sharp images (before interpolation) to synthesize one blurry image.

\noindent{\bf HQF.} The HQF dataset \cite{stoffregen2020reducing} contains real events and sharp clear video frames captured simultaneously by a DAVIS240 camera. The motion blur is synthesized following the same approach as the REDS dataset, where we first increase the frame rate of the captured sharp videos from 25 fps to 200 fps and then average 49 consecutive frames to generate blurry images. Thus each blurry image corresponds to 7 sharp frames in the original captured videos, which are defined as the ground truths. 
% We use the same approach as the REDS dataset (interpolation and averaging) to synthesize motion blur, where 7 sharp images are merged into one blurry image for the HQF dataset. 

\noindent{\bf RBE.} The RBE dataset~\cite{xu2021motion} employs a DAVIS346 camera to collect real-world blurry videos and the corresponding event streams, without ground-truth sharp images. Therefore, we use it to validate the effectiveness of our method in real-world scenarios.

\begin{table}[t]
    \centering
    \caption{
    Quantitative comparisons for continuous-time restoration on the REDS dataset where 31 images are restored for each blurry frame. {\it Use events} indicate if the deblurring method uses events.
    %{\it Use events} indicate if a method uses
    % events and {\it Interpolation} indicate if an additional Video Frame Interpolation (VFI) method is cascaded to achieve continuous-time restoration. 
    }
    \renewcommand\arraystretch{1.2}
    \label{tab:reds_31}
    \begin{tabular}{lccccc}
        \hline
        Methods & Use events & PSNR$\uparrow$ & SSIM$\uparrow$ \\
        \hline
        Motion-ETR \cite{zhang2021exposure} & \ding{56}  & 19.084 & 0.4857 \\
        EDI \cite{pan2019bringing}   & \ding{52}  & 20.783 & 0.6179 \\
        eSL-Net \cite{wang2020event} & \ding{52}  & 21.166 & 0.6779 \\
        E-CIR \cite{song2022cir}     & \ding{52}  & 25.462 & 0.7919 \\
        LEVS \cite{jin2018learning} + Timelens \cite{tulyakov2021time} & \ding{56} & 19.981 & 0.5359 \\
        LEDVDI \cite{lin2020learning}  + Timelens \cite{tulyakov2021time} & \ding{52}  & 27.271 & 0.8508 \\
        RED-Net \cite{xu2021motion}  + Timelens \cite{tulyakov2021time} & \ding{52}  & 28.051 & 0.8686 \\
        \hline 
        Ours                         & \ding{52}  & \bf{30.576} & \bf{0.9173}\\ 
        \hline
    \end{tabular}
    % \vspace{-1em}
\end{table}

\subsubsection{Implementation Details}
The network is implemented using Pytorch and trained on two NVIDIA GeForce RTX 3090 GPUs.
% with batch size 8 by default. 
The training process is composed of two phases. In each phase, we utilize a batch size of 8 and employ the Adam optimizer \cite{kingma2014adam} with momentum and momentum2 as 0.9 and 0.999.
% The learning rate starts at $3 \times 10^{-4}$ and decays from $25$th to $100$th epoch to $1\times 10^{-5}$,  
We randomly crop the images to $128 \times 128$ patches and apply horizontal flipping for data augmentation.
% The training dataset is augmented by randomly flipping the input images and cropping them into $128 \times 128$ patches.

During the first phase, the IVF network is trained for 400 epochs where the learning rate is $1 \times 10^{-4}$ in the first 50 epochs, linearly decays to $1\times 10^{-5}$ until the $200$-th epoch, and remains unchanged to the end. We set the weighting factors $\lambda_1=1$ and $\lambda_2=0$ in the initial $300$ epochs to stabilize the reconstruction quality of the latent frames at reference timestamps and then modify them to $\lambda_1=0.2$ and $\lambda_2=1$ for the rest 100 epochs to supervise the restoration at arbitrary time instances. For the motion-guided loss $\mathcal{L}_{motion}$, we set the parameters $M=3$ and $N=2$ and employ the optical flow predicted by the EV-Flow network \cite{zhu2018ev} which is pre-trained on the MVSEC dataset~\cite{zhu2018ev} and then fine-tuned during the training stage.

In the second phase, the EER model is trained individually for 200 epochs with the learning rate initialized as $1 \times 10^{-4}$ and decayed by 0.8 every 50 epochs.
% The EV-Flow network \cite{zhu2018ev} is employed for the flow loss $\mathcal{L}_{flow}$ is pretrained on the MVSEC dataset~\cite{zhu2018ev} and fine-tuned with our network in the training stage. 
All input blurry images and the concurrent events are temporally and spatially aligned  before feeding into the network. For an event stream, we first divide it into $6$ segments with equal time intervals and then convert each segment into the accumulated event frame and time surface \cite{zhu2018ev} as the input.

\par 
Our proposed DeMo-IVF is trained on a joint training set from both REDS and HQF, where we set the normalized timestamps $\mathcal{T}_K\triangleq [ 0, \frac{1}{6},...,\frac{5}{6},1 ]$ as the referenced timestamps and only use the corresponding 7 ground-truth images for supervision. Then the performance of the single-frame and the video-sequence restorations are evaluated at the referenced timestamps. For the REDS dataset, we leave the remaining $24$ ground-truth latent images corresponding to the non-referenced timestamps, which enable the evaluation of reconstruction at non-referenced timestamps on the REDS dataset. 

\subsection{Comparisons with State-of-the-Art Methods}\label{sec:6-2}
% 比较了哪些算法
In this subsection, we compare our method with state-of-the-art image-only and event-based deblurring methods capable of recovering sharp image sequences from a single blurry image. The image-only methods include LEVS\cite{jin2018learning} and Motion-ETR\cite{zhang2021exposure}, and the event-based methods include EDI\cite{pan2019bringing}, eSL-Net\cite{wang2020event}, LEDVDI\cite{lin2020learning}, RED-Net\cite{xu2021motion} and E-CIR\cite{song2022cir}. 
LEVS, LEDVDI, and RED-Net can convert one blurry image into a sharp video sequence composed of 7 frames, while Motion-ETR, EDI, eSL-Net, and E-CIR can restore latent sharp images at arbitrary timestamps as our method. For the sake of fair comparisons, eSL-Net, LEDVDI, RED-Net, and E-CIR are all fine-tuned on the training sets of the REDS and HQF datasets with the supervision of 7 ground-truth images at the referenced timestamps. 
% Existing motion deblurring approaches mainly focus on converting one blurry image into a single latent image or a video sequence, while our proposed DeMo aims to recover latent sharp images at arbitrary time instances. Thus, we conduct experiments on both setting.

\subsubsection{Single Frame and Video Sequence Restoration}
\label{comparison}
% \noindent {\bf }
We first evaluate the performance of all methods on the REDS and HQF datasets for conventional motion deblurring tasks from a single blurry image, \ie, restoring a single sharp image or a video sequence with 7 sharp images. 
In this experiment, only the sharp images located at referenced timestamps are used for evaluation.
% For the synthetic REDS dataset, one blurry image corresponds to 31 sharp images, and only 7 images located in specific timestamps are used for evaluation.
The quantitative results are presented in \cref{tab:sota}. As we can see, on the REDS dataset, the proposed method outperforms other methods in terms of both PSNR (up to 1.68 dB improvement) and SSIM (up to 0.0376 improvements). 
% These improvements stem from using the dual attention mechanism (DAM), which allows for a mutual enhancement between the events and the blurry image.
% On the other hand, we also validate the superior performance of DeMo with real-world events on the HQF dataset. 
On the HQF dataset, our method still performs the best, which also validates the effectiveness of our model in handling real-world events.

%Note that the inconsistency between the synthesized and real-world events brings a gap to real-world scenarios for most learning based algorithms, including DeMo-$7$.  Comparing to the methods trained with real events, \ie, RED-Net and LEDVDI,  our proposed DeMo-$7$ still exhibits better performance in terms of both quantitative and qualitative results (as shown in Fig.~\ref{fig:indoor}), which validates the generalization of our proposed network.
% Thus, the performance drop is considerable for both IVF-$7$ and eSL-Net which only see synthetic events.
% while RED-Net and LEDVDI are trained with real events to reduce the gap to real scenes.
% For the methods trained only with synthetic events, both IVF-$7$ and eSL-Net exhibit a considerable decrease on PSNR since they never see real events.
% It is shown that IVF-$7$ can still achieve the best performance in terms of both PSNR and SSIM. 
% event-based approaches exhibit overwhelming performance deterioration due to the data inconsistency, especially for learning based approaches trained only with synthetic events, \eg, eSL-Net. 
% Since the HQF dataset is built with real events and simulated motion blurs, most algorithms exhibit a large gap between REDS and HQF, for instance, eSL-Net. Although our method is only trained with synthetic events on the REDS dataset without seeing any real events, it still performs well on HQF and exceeds the performance of state-of-the-art methods. Note that the RED-Net and LEDVDI are all trained with real events and adapt to the real scenes, which validates the generalization of our proposed IVF. 

Correspondingly, we demonstrate a qualitative comparison in \cref{fig:reds_hqf_7}, where we select two exemplar restorations respectively from the REDS and HQF datasets. 
Due to the inherent ambiguities of temporal ordering and lost spatial textures in the blurry image, LEVS and Motion-ETR fail to recover the latent sharp images without the aid of events. 
Especially for severe blur caused by high-speed motion, the image-based methods are unable to recover the hidden moving targets such as the {\it walking woman} of the first example in \cref{fig:reds_hqf_7}.
Event-based approaches outperform image-based methods with significant improvements thanks to the introduction of events. 
% 基于事件的方法
% In contrast, event-based methods can achieve better performance than image-based methods. 
EDI can reconstruct the motion target, but the details are still blurry, possibly due to the fact that EDI computes the double integral for each pixel independently and does not take full advantage of the spatially structured information provided by events.
% possibly caused by the accumulation of event noise assumption of constant triggering thresholds. 
The learning-based methods significantly improves the performance of event-based motion deblurring by learning convolutional neural networks from large volumes of data, \eg, eSL-Net, E-CIR, LEDVDI, RED-Net, and our proposed DeMo-IVF.
Nevertheless, eSL-Net tends to produce halo effects along the high-contrast edges caused by modeling errors introduced when reversing the event polarity in the reshuffle process, while E-CIR would suffer from significant noise artifacts induced by input event noise. Compared to eSL-Net and E-CIR, LEDVDI and RED-Net predict relatively sharper images, but distortions and blurriness still exist in the foreground targets and background textures, \eg, the white stripe on the clothes of the first example and the gaps between the floor tiles of the second example.
% avoids this problem by recovering continuous intensity with parametric polynomials, the event noise , the reconstruction images present some noticeable noise artifacts.
Our proposed DeMo-IVF gives results with sharper edges and smoother surfaces than the state-of-the-art methods, demonstrating the superiority of our dual attention mechanism in the DFEN, where mutual compensation between the events and the blurry image is achieved. Meanwhile, motion- and texture-guided supervisions provide both temporal consistencies and texture enhancements over the restored video sequences as shown in \cref{fig:reds_31} (restorations of referenced timestamps), which significantly improves the deblurring performance with smooth inter-frame transitions and sharp texture edges.

\subsubsection{Continuous-Time Video Restoration}
\label{sec:continuous-time video restoration}
To explore the superiority of DeMo-IVF to recover latent sharp images at arbitrary timestamps, we conduct experiments on reconstructing all 31 frames on the REDS dataset corresponding to 7 referenced timestamps and 24 non-referenced timestamps. Besides one-stage methods, \ie, Motion-ETR~\cite{zhang2021exposure}, EDI~\cite{pan2019bringing}, eSL-Net~\cite{wang2020event}, and E-CIR~\cite{song2022cir}, we also compared our proposed method to two-stage methods by cascading sequence deblurring approaches, \ie, LEVS~\cite{jin2018learning}, LEDVDI~\cite{jiang2020learning}, and RED-Net~\cite{xu2021motion}, and the event-based video frame interpolation method, \ie, Timelens~\cite{tulyakov2021time} to achieve continuous-time restoration, where 7 deblurred images are interpolated to 31 images.

The quantitative and qualitative results are presented in \cref{tab:reds_31} and \cref{fig:reds_31} respectively. Our proposed DeMo-IVF outperforms both one-stage and two-stage approaches by a large margin. Considering the two-stage methods, LEDVDI+Timelens and RED-Net+Timelens even have a large performance drop compared to reconstructing only 7 frames since the deblurring errors might be propagated to the interpolation stage and accumulated to the restorations of the non-referenced timestamps, while LEVS+Timelens performs with a slight improvement compared to LEVS on sequence restorations since the introduction of events in the interpolation stage.
% The results of LEVS do not change much, probably because the recovered images are too blurry and the effect of the interpolation is attenuated. 
For the one-stage methods, event-based approaches perform much better than the image-only approach, \ie, Motion-ETR, since events can provide the intra-frame information in terms of motions and textures. Among the event-based approaches, the learning-based approaches perform better than the optimization-based method, \ie, EDI, while the performance eSL-Net is confined by initial deblurring results as shown in \cref{fig:reds_hqf_7} and E-CIR suffers from the event noise problem as shown in \cref{fig:reds_31}. Compared to the event-based approaches, our proposed DeMo-IVF restores sharper and clearer latent images with high-contrast textures and smooth inter-frame transitions, thus gaining quantitatively higher PSNR and SSIM.

% especially in the learning-based methods. This may be because most learning-based approaches require the representation of events in the form of frames, which is usually achieved by accumulating events over a small period of time, and information lost in the process affects high frame-rate image recovery. To retain more spatio-temporal information in events, our approach chooses the representation of the time surface combined with cumulative events and complements extra event frames around input timestamps in the EER module, but it still cannot avoid performance deterioration.

% Moreover, we see that the existing one-stage methods perform worse than the two-stage methods. 
% However, in general, our method still has the best performance compared to the one- and two-stage methods. As shown in \cref{fig:reds_31}, the sharp images reconstructed by our method own finer texture and detail, while the image sequence is more consistent and smoother.
% RBE数据集上的结果

To further demonstrate the generalizability of our proposed DeMo-IVF in real-world scenarios, we also perform continuous-time video reconstruction over the RBE dataset with real events and blurry images. Without ground truth, only qualitative comparisons can be made as shown in \cref{fig:rbe} and we obtain consistent performance as on synthetic REDS and HQF datasets. Specifically, the restored {\it cube} and {\it chessboard} by our proposed DeMo-IVF are apparently with higher quality than that restored by E-CIR and RED-Net, which validates the generalizability of our method.

\begin{table}[t]
\caption{
Ablation study of the dual feature path, DAM, MLP and EER in our method on the REDS dataset. All the models are trained using the same strategy.
% We train all these models with the same setting as DeMo-IVF.
}
\centering
\renewcommand\arraystretch{1.2}
\begin{tabular}{c|cccc|cc}
% \begin{tabular}{p{0.5cm}<{\centering}|p{1.1cm}<{\centering} p{1.3cm}<{\centering} p{1.1cm}<{\centering}|p{1.25cm}<{\centering} p{1.25cm}<{\centering}}
\hline
Ex. & Dual      & DAM       & MLP       & EER       & PSNR$\uparrow$    & SSIM$\uparrow$   \\ \hline
1   &           &           & \ding{52} & \ding{52} & 30.286            & 0.8822 \\
2   &\ding{52}  &           & \ding{52} &           & 29.203            & 0.8583 \\
3   &\ding{52}  &           & \ding{52} & \ding{52} & 30.160            & 0.8817 \\
4   &\ding{52}  & \ding{52} &           & \ding{52} & 30.951            & 0.8968 \\
5   &\ding{52}  & \ding{52} & \ding{52} &           & 30.182            & 0.8814 \\
6   &\ding{52}  & \ding{52} & \ding{52} & \ding{52} & {\bf 31.111} & {\bf 0.8995} \\ \hline
% 5   &\checkmark & \checkmark &           &\checkmark& 29.842  & 28.463 \\
% 6   &\checkmark & \checkmark &\checkmark &          & 30.023  & 28.874\\
% 7&\checkmark & \checkmark &\checkmark&\checkmark&                       & - & - \\
\end{tabular}
\label{tab:ablation}
% \vspace{-1.5em}
\end{table}

\subsection{Ablation Study}\label{sec:6-3}

In this subsection, we present ablation studies to analyze the design choices of our method. We first demonstrate the performance contribution of each module in the network architecture (\cref{tab:ablation}). Then, we further validate the effectiveness of DAM with additional noise experiments (\cref{tab:noise} and \cref{fig:noise}). Finally, we analyze the role of each supervision in the training strategy (\cref{tab:reds_flow_er} and \cref{fig:ablation_flow_er}).

\subsubsection{Network Architecture} 
% The proposed method learns the scene dynamics as the IVF from a single blurry image and the concurrent event, which enables the latent image restoration of arbitrary timestamps. It has been shown that the proposed IVF performs favorably against state-of-the-art methods. 
The proposed network architecture is composed of a dual feature embedding network, \ie, DFEN, where the dual feature path and the DAM are designed, a continuous-time decoder MLP, and an event edge refinement (EER) module.
The ablation studies are conducted on the synthetic REDS dataset, where the sequence restoration task is considered and 6 different experiments are implemented to analyze the effectiveness of each component, as shown in \cref{tab:ablation}.

First, we remove the DAM and replace it with two other fusion mechanisms: (Ex. 1) a single feature path that directly receives the concatenated blurry image and the corresponding event frames; (Ex. 3) a dual feature path that concatenates after extracting features from blurry image and event frames separately. These two methods can be regarded as the pre-fusion and post-fusion respectively. Compared to them, our method utilizes the DAM to take the complementary merits of frames and events, where a bidirectional enhancement process is implemented to suppress the noise in event features and simultaneously deblur image features. And thus we can find that such fusion mechanism improves the performance by a large margin (0.825 dB in PSNR and 0.0173 in SSIM). Moreover, we find that pre-fusion by concatenation of two modality sources performs better than the model with only a dual feature path, where the mutual compensation of dual features is not fully explored. This further proves the importance of DAM in our DFEN in terms of motion deblurring. In our method, the continuous-time decoding MLP is the crucial module to learn the implicit video function. The deletion of this module (Ex. 4) makes our method degenerate into a sequence restoration model and slightly reduces the performance (0.16 dB in PSNR and 0.0028 in SSIM). The EER module is designed to supplement the lost information among the event frames and refine the detailed textures. The comparison between Ex. 5 and Ex. 6 (or Ex. 2 and Ex. 3) validates its effectiveness.

\subsubsection{Noise Suppression} 
To further verify the effectiveness of DAM in noise suppression, we evaluate our model (with DAM) and the model without DAM, \ie, Ex. 3 in \cref{tab:ablation}, on the REDS dataset that is contaminated by noisy events generated from the uniform random distribution as \cite{wang2020event}. 
In \cref{tab:noise}, we compare the performance of the above two models under different noise levels ranging from 0\% to 30\%, where the noise level is defined as the proportion of noisy events to the original events. In the cases of low-level event noise (0\% and 5\%), the model with DAM already outperforms its counterpart by a large margin, benefiting from the mutual compensation of image and event features in DAM. As the noise level rises, the performance gap becomes more evident since our network can exploit the smooth image features to enhance the information in event features while suppressing noise.  

% showing that DAM plays a role in noise suppression.

% As shown in \cref{tab:noise}, we evaluate the performance changes of both networks under different level of event noise.
% Event-based deblurri  ng method relies on high-quality events, naturally, with the increasing proportion of event noise, the process of deblurring will be seriously affected and the performance shows a downward trend for both networks.
% However, it can see that the trend of network performance decreasing with increasing noise ratio becomes slower for our network comparing to another one, which shows that DAM plays a role in noise suppression.

% The dual attention mechanism is used for mutual compensation between the blurry image and events, including a bidirectional enhancement, \ie, E2B and B2E. 
% To validate their effectiveness, four experiments (Ex. 4-6 and 8) have been carried out. It is shown that B2E and E2B respectively bring PSNR gain of 0.58 dB and {0.861 dB}. While combining B2E and E2B can achieve further improvement (1.014 dB in PSNR) which validates the importance of the dual attention mechanism.

% \par
% \noindent\textbf{MLP.} In DeMo, MLP is exploited to learn the implicit video function. We can observe from Ex. 7 that the MLP layer can improve the performance by 0.292 dB in PSNR.

In addition to quantitative comparisons, the qualitative results are depicted in \cref{fig:noise}.
In the noise-free scenarios, the networks with and without DAM are both able to remove motion blurs and restore sharp images by utilizing high-quality event streams. However, with the surge in the noise level, the model without DAM is severely disturbed by event noise, leading to noisy textures and unpleasant visual effects.
% It can be seen that, without adding event noise, both networks can utilize the high-quality event stream to remove motion blur and restore sharp images.
% With the increased proportion of event noise, the network without DAM is severely affected by that and produce outputs with much noise and artifacts. 
By utilizing the less noisy image features, DFEN mitigates the disturbance and produces smoother and more realistic textures, which further illustrates the importance of DAM for event noise suppression.

\begin{table*}[!ht]
    \centering
    \caption{
    Ablation study of IVF model with (w) and without (w/o) DAM on the REDS dataset with 0\%, 5\%, 20\%, and 30\% event noise.
    The last row (Gain) shows the performance improvement by DAM.
    }
    \renewcommand\arraystretch{1.2}
    \begin{tabular}
    {cccccccccccc}
    % {c|p{1.2cm}<{\centering}p{1.2cm}<{\centering}|p{1.2cm}<{\centering}p{1.2cm}<{\centering}|p{1.2cm}<{\centering}p{1.2cm}<{\centering}|p{1.2cm}<{\centering}p{1.2cm}<{\centering}}
    \hline
    \multirow{2}{*}{Methods} 
    & \multicolumn{2}{c}{0\%} & & \multicolumn{2}{c}{5\%} & & \multicolumn{2}{c}{20\%} & & \multicolumn{2}{c}{30\%} \\ 
    \cline{2-3} \cline{5-6} \cline{8-9} \cline{11-12}
    & PSNR$\uparrow$ & SSIM$\uparrow$ & & PSNR$\uparrow$ & SSIM$\uparrow$ & & PSNR$\uparrow$ & SSIM$\uparrow$ & & PSNR$\uparrow$ & SSIM$\uparrow$ \\ 
     \hline
    w/o DAM   & 29.203 & 0.8583 &   & 28.546 & 0.8364 &   & 26.933 & 0.7567 &   & 26.089 & 0.7141     \\ 
    w DAM     & {\bf 30.182} & {\bf 0.8814} &   & {\bf 29.390} & {\bf 0.8612}&   & {\bf 28.161} & {\bf 0.8191} &    & {\bf 27.477} & {\bf 0.7916}     \\ 
    \hline
    Gain      & 0.979  & 0.0231 &   & 0.844  & 0.0248 &   & 1.228  & 0.0624 &    & 1.388 & 0.0775     \\ 
    \hline
    \end{tabular}
    \label{tab:noise}
\end{table*}
\begin{table}[t]
    \centering
    \caption{Ablation study of the training strategy over the REDS dataset.}
    \renewcommand\arraystretch{1.2}
    \begin{tabular}{c | ccc|cc}
        \hline
        Ex. & $\mathcal{L}_{im}$ & $\mathcal{L}_{motion}$ & $\mathcal{L}_{texture}$ & PSNR$\uparrow$ & SSIM$\uparrow$ \\
        \hline
        A   & \ding{52} &           &           & 26.461 & 0.7955 \\
        B   & \ding{52} &           & \ding{52} & 27.259 & 0.8217 \\
        C   & \ding{52} & \ding{52} &           & 29.416 & 0.8953 \\
        D   & \ding{52} & \ding{52} & \ding{52} & 30.576 & 0.9173 \\
        % E   & \ding{52} & \ding{52} &           & \ding{52} & 30.708 & 0.9160 \\
        \hline 
        % Ours                    & \ding{52}  & \ding{56} & \bf{29.855} & \bf{0.9023}\\ 
        % \hline
    \end{tabular}
    \label{tab:reds_flow_er}
\end{table}

\input{fig/fig_noise}
\input{fig/fig_flow_er.tex}
\subsubsection{Training Strategy}
% To alleviate the bias issue caused by sparse sampling, we apply the flow loss $\mathcal{L}_{flow}$ to supervise the restoration of frames without references. 
Our training process is divided into two stages, \ie, optimizing DFEN and MLP with $\mathcal{L}_{im}$ and $\mathcal{L}_{motion}$, and optimizing the EER module with $\mathcal{L}_{texture}$. In this section, we present investigations on the ablation of the loss functions in each step.
% For training strategy, $L_{im}$, $L_{motion}$ and $L_{EER}$ are used to optimize the model. To verify their effects, the ablation experiments is conducted on the REDS dataset. 
The results are presented respectively in \cref{tab:reds_flow_er} and \cref{fig:ablation_flow_er}. 
% 分析实验结果

% A 仅使用key监督训练的结果
% \noindent{\bf Supervision only at Referenced Timestamps.}

% B 同时使用key监督和flow loss的结果
\noindent{\bf Motion-Guided Supervision.} Only supervised with the $\mathcal{L}_{im}$ loss (Ex. A), the model only reconstructs sharp images at the referenced timestamps but performs poorly at non-referenced timestamps, where its reconstructions exhibit severe ghosting artifacts with overlays of the two adjacent sharp frames of reference timestamps. It indicates that the sparse sampling in $\mathcal{L}_{im}$ loss will bias the network towards learning to reconstruct sharp images at the timestamps with references. Therefore, motion-guided supervision is introduced to tackle this problem. When the $\mathcal{L}_{motion}$ loss is included (Ex. C and Ex. D), the network learns to produce temporally smooth inter-frame transitions of the restored video sequence, alleviating the bias issue. The resulting model can predict sharp images at both reference and non-reference timestamps as shown in \cref{fig:ablation_flow_er} and achieves a large performance improvement (2.955 dB in PSNR and 0.0998 in SSIM between Ex. A and Ex. C, 3.317 dB in PSNR and 0.0956 in SSIM between Ex. B and Ex. D) as shown in \cref{tab:reds_flow_er}. 
% Nevertheless, we can observe that the restored images of the referenced timestamps are noisier and less sharp than (Ex. A) only with $\mathcal{L}_{im}$, as shown in \cref{fig:ablation_flow_er}. Such defects might be induced by the error of the predicted optical flow.
% The reconstruction quality on reference timestamp even drops compared to Ex.A due to the error introduced by the estimated optical flow.

% C EER
\noindent{\bf Texture-Guided Supervision.} Only with the motion-guided supervision, we can observe that the restored images of the referenced timestamps are noisier and blurrier than (Ex. A) only with $\mathcal{L}_{im}$, as shown in \cref{fig:ablation_flow_er}. Such defects might be induced by the error of the predicted optical flow. To tackle this problem, texture-guided supervision is introduced by the EER module and $\mathcal{L}_{texture}$, enhancing the sharpness of restorations. As we further add the EER module (Ex. D), the performance is significantly improved (1.16 dB in PSNR and 0.022 in SSIM) and the resulting model can generate sharper restorations and higher contrast textures. To further validate the effectiveness of texture-guided supervision, we add EER on the basis of Ex. A and train the model Ex. B with $\mathcal{L}_{texture}$. One can observe that the introduction of texture-guided supervision can also bring noticeable performance improvement (0.798 dB in PSNR and 0.0262 in SSIM), which validates its effectiveness. 
\section{Conclusion}
This paper proposes a novel DeMo-IVF method to fully demystify motion-blurred observations by learning an implicit video function from a single blurry image and the concurrent event streams. 
Different from existing deblurring methods, our proposed DeMo-IVF is able to query the latent sharp images at arbitrary timestamps within the exposure period of the blurry input. 
Specifically, a dual-feature embedding network is proposed to make full use of frames and events, simultaneously achieving blurry feature enhancement and event noise suppression, while an event-based edge refinement module is presented to enhance the texture restoration performance.
Based on limited ground-truth images of referenced timestamps, the motion- and texture-guided supervisions are further utilized to train the overall network.
Extensive experiments on both synthetic and real-world datasets demonstrate that our DeMo-IVF achieves state-of-the-art deblurring performance and fully recovers the scene dynamics behind blurry images.

\ifCLASSOPTIONcaptionsoff
  \newpage
\fi



%\bibliographystyle{IEEEtran}
\bibliographystyle{IEEEtran}
\bibliography{IEEEabrv,main}

\begin{IEEEbiography}[{\includegraphics[width=1in,height=1.25in,clip,keepaspectratio, trim={20 0 20 60}]{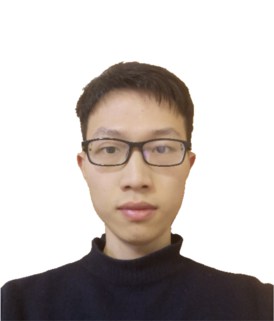}}]{Zhangyi Cheng}
received his B.E. degree in computer science and technology from Wuhan University, Wuhan, China, in 2021. He is currently working toward an M.S. degree in artificial intelligence with the school of computer science, Wuhan University, Wuhan, China. His research interests include computer vision and neuromorphic computation.
\end{IEEEbiography}

% \vskip -2.3\baselineskip plus -1fil
% \vskip -2.\baselineskip plus -1fil

\begin{IEEEbiography}[{\includegraphics[width=1in,height=1.25in,clip,keepaspectratio, trim={20 40 20 40}]{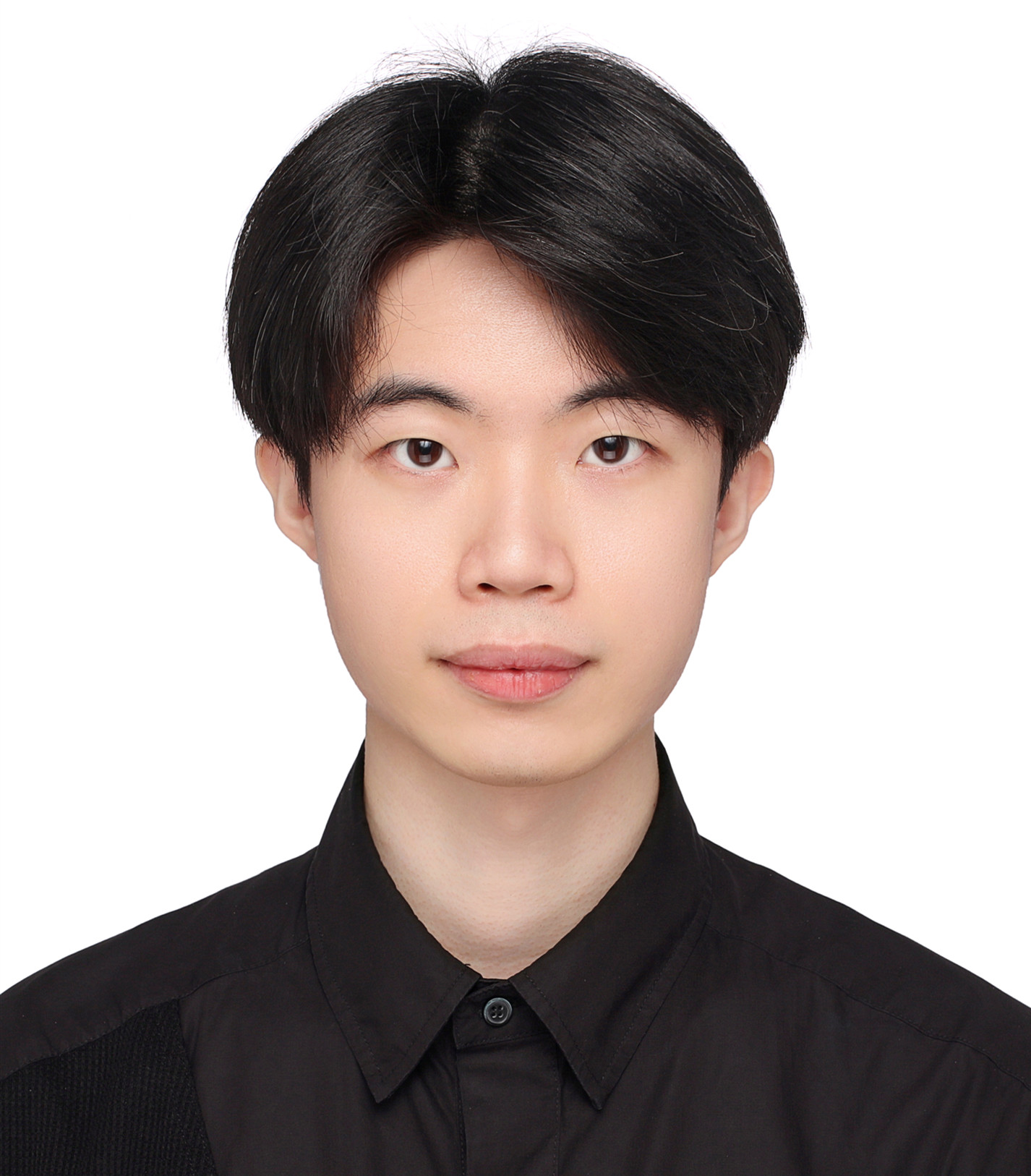}}]{Xiang Zhang}
received his B.E. degree in communication engineering from Wuhan University, Wuhan, China, in 2020. He is currently working toward an M.S. degree in information and communication engineering with the electronic information school, Wuhan University, Wuhan, China. His research interests include computer vision and neuromorphic computation.
\end{IEEEbiography}

% \vskip -2\baselineskip plus -1fil

\begin{IEEEbiography}[{\includegraphics[width=1in,height=1.25in,clip, keepaspectratio, trim={0 50 0 90}]{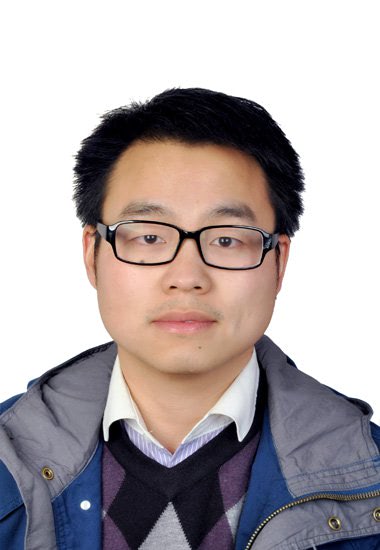}}]{Lei Yu}
received his B.S. and Ph.D. degrees in signal processing from Wuhan University, Wuhan, China, in 2006 and 2012, respectively. From 2013 to 2014, he has been a Postdoc Researcher with the VisAGeS Group at the Institut National de Recherche en Informatique et en Automatique (INRIA) for one and a half years. He is currently working as an associate professor at the School of Electronics and Information, Wuhan University, Wuhan, China. From 2016 to 2017, he has also been a Visiting Professor at Duke University for one year. He has been working as a guest professor in the École Nationale Supérieure de l'Électronique et de ses Applications (ENSEA), Cergy, France, for one month in 2018. His research interests include neuromorphic vision and computation.
\end{IEEEbiography}

% \vskip -2.3\baselineskip plus -1fil
% \vskip -2.\baselineskip plus -1fil

\begin{IEEEbiography}[{\includegraphics[width=1in,height=1.25in,clip,keepaspectratio]{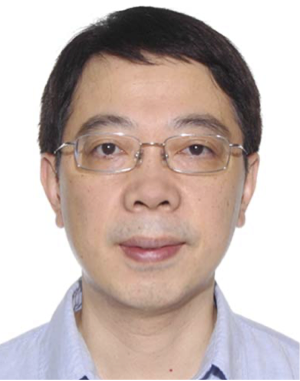}}]{Jianzhuang Liu}
received the Ph.D. degree in computer vision from The Chinese University of Hong Kong, Hong Kong, in 1997. From 1998 to 2000, he was a Research Fellow with Nanyang Technological University, Singapore. From 2000 to 2012, he was a Postdoctoral Fellow, an Assistant Professor, and an Adjunct Associate Professor with The Chinese University of Hong Kong. In 2011, he joined the Shenzhen Institute of Advanced Technology, University of Chinese Academy of Sciences, Shenzhen, China,
as a Professor. He is currently a Principal Researcher with Huawei Technologies Company Ltd., Shenzhen. He has authored more than 150 papers. His research interests include computer vision, image processing, deep learning, and graphics.
\end{IEEEbiography}

% \vskip -2.3\baselineskip plus -1fil
% \vskip -2.\baselineskip plus -1fil

\begin{IEEEbiography}[{\includegraphics[width=1in,height=1.25in,clip,keepaspectratio, trim={0 35 0 10}]{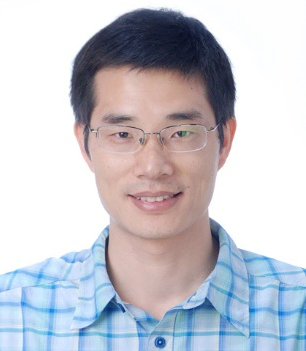}}]{Wen Yang} received the B.S. degree in electronic apparatus and surveying technology, and the M.S. degree in computer application technology and the Ph.D. degree in communication and information system from Wuhan University, Wuhan, China, in 1998, 2001, and 2004, respectively. From 2008 to 2009, he worked as a Visiting Scholar with the Apprentissage et Interfaces (AI) Team, Laboratoire Jean Kuntzmann, Grenoble, France. From 2010 to 2013, he worked as a Post-Doctoral Researcher with the State Key Laboratory of Information Engineering, Surveying, Mapping and Remote Sensing, Wuhan University. Since then, he has been a Full Professor with the School of Electronic Information, Wuhan University. He is also a guest professor of the Future Lab AI4EO in Technical University of Munich. His research interests include object detection and recognition, multisensor information fusion, and remote sensing image processing.
\end{IEEEbiography}

% \vskip -2.3\baselineskip plus -1fil
% \vskip -2.\baselineskip plus -1fil

\begin{IEEEbiography}[{\includegraphics[width=1in,height=1.25in,clip,keepaspectratio]{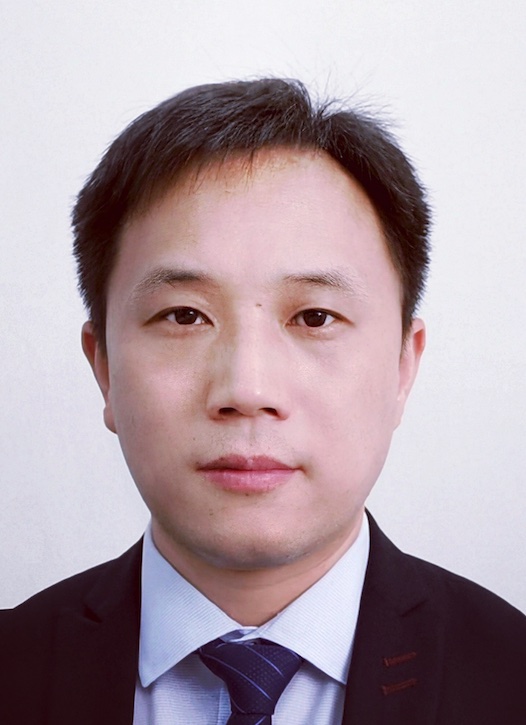}}]{Gui-Song Xia}
received his Ph.D. degree in image processing and computer vision from CNRS LTCI, T{\'e}l{\'e}com ParisTech, Paris, France, in 2011. From 2011 to 2012, he has been a Post-Doctoral Researcher with the Centre de Recherche en Math{\'e}matiques de la Decision, CNRS, Paris-Dauphine University, Paris, for one and a half years.
He is currently working as a full professor in computer vision and photogrammetry at Wuhan University. He has also been working as Visiting Scholar at DMA, {\'E}cole Normale Sup{\'e}rieure (ENS-Paris) for two months in 2018. He is also a guest professor of the Future Lab AI4EO in Technical University of Munich (TUM). His current research interests include mathematical modeling of images and videos, structure from motion, perceptual grouping, and remote sensing image understanding. He serves on the Editorial Boards of several journals, including {\em ISPRS Journal of Photogrammetry and Remote Sensing, Pattern Recognition, Signal Processing: Image Communications, EURASIP Journal on Image \& Video Processing, Journal of Remote Sensing, and Frontiers in Computer Science: Computer Vision}.
\end{IEEEbiography}

\end{document}